\documentclass[9.5pt,journal,finalsubmission,compsoc]{IEEEtran}
\ifCLASSOPTIONcompsoc
  \usepackage[nocompress]{cite}
\else
  \usepackage{cite}
\fi
	\usepackage[pdftex]{graphicx}
   \DeclareGraphicsExtensions{.pdf,.jpeg,.png}
\usepackage[cmex10]{amsmath}

\ifCLASSOPTIONcompsoc
  \usepackage[caption=false,font=normalsize,labelfont=sf,textfont=sf]{subfig}
\else
  \usepackage[caption=false,font=footnotesize]{subfig}
\fi

\begin{document}
%
\title{A General Methodology for the Determination of 2D Bodies Elastic Deformation Invariants. Application to the Automatic Identification of Parasites}
%
%
%
%

\author{Dimitris~Arabadjis,~
        Panayiotis~Rousopoulos,~
        Constantin~Papaodysseus,~
        Michalis~Panagopoulos,~
        Panayiota~Loumou~
        and~Georgios~Theodoropoulos,
\IEEEcompsocitemizethanks{\IEEEcompsocthanksitem D. Arabadjis, P. Rousopoulos, C. Papaodysseus, M. Panagopoulos, P. Loumou are with the National Technical University of Athens, School of Electrical and Computer Engineering, GR-15773 Athens,Greece (telephone: +30210 7722329 e-mail: cpapaod@ cs.ntua.gr)
\IEEEcompsocthanksitem G. Theodoropoulos is with the Agricaltural University of Athens}
\thanks{\copyright ~ 2010 IEEE. Personal use of this material is permitted. Permission from IEEE must be obtained for all other uses, in any current or future media, including reprinting/republishing this material for advertising or promotional purposes, creating new collective works, for resale or redistribution to servers or lists, or reuse of any copyrighted component of this work in other works.}
}

\IEEEcompsoctitleabstractindextext{%
\begin{abstract}
A novel methodology is introduced here that exploits 2D images of arbitrary elastic body deformation instances, so as to quantify mechano-elastic characteristics that are deformation invariant. Determination of such characteristics allows for developing methods offering an image of the undeformed body. General assumptions about the mechano-elastic properties of the bodies are stated, which lead to two different approaches for obtaining bodies' deformation invariants. One was developed to spot deformed body's neutral line and its cross sections, while the other solves deformation PDEs by performing a set of equivalent image operations on the deformed body images. Both these processes may furnish a body undeformed version from its deformed image. This was confirmed by obtaining the undeformed shape of deformed parasites, cells (protozoa), fibers and human lips. In addition, the method has been applied to the important problem of parasite automatic classification from their microscopic images. To achieve this, we first apply the previous method to straighten the highly deformed parasites and then we apply a dedicated curve classification method to the straightened parasite contours. It is demonstrated that essentially different deformations of the same parasite give rise to practically the same undeformed shape, thus confirming the consistency of the introduced methodology. Finally, the developed pattern recognition method classifies the unwrapped parasites into 6 families, with an accuracy rate of 97.6 \%.
\end{abstract}

}

\maketitle

\IEEEdisplaynotcompsoctitleabstractindextext

%
\IEEEpeerreviewmaketitle

\section{Introduction}
%
%

%
%
%
%
\IEEEPARstart{T}{here} are numerous applications, where bodies suffer deformation due to elastic forces (stresses). In these cases, one frequently encounters two important problems: a) to make consistent and reliable estimation of the body undeformed shape from images of random instances of body deformation and b) to identify the deformed body from these images. We would like to emphasize that, as a rule, identification of bodies on the basis of images of their deformation, is practically prohibited by the randomness of the deformation. One encounters such problems in various disciplines applications, such as automatic identification of highly deformed parasites, cells or large molecules from their images obtained via microscope, like the problem tackled in \cite{bib16}, in strength of materials, elastography \cite{bib20}, \cite{bib9}, in civil engineering in general, etc. 

In the bibliography, there are some approaches for generating the phases between an initial and a final stage of a body elastic deformation \cite{bib14}, \cite{bib19}. In these publications, images of both the initial and final stages deformation are available. On the contrary, in the present paper, images of the undeformed body are not a priori available, hence there are no initial phase representations. Moreover, methods that do not demand given initial body image to perform deformation, use deformable lines or surfaces \cite{bib4}, \cite{bib17}, which are not uniquely connected to deformation invariants of body shape. Thus, they cannot be used to create representative undeformed body images from its deformed instances. There are, also, some approaches dealing with deformable contours in the concept of their invariant formulation \cite{bib6}, \cite{bib7}. In these references, active contour models are formulated so as their body shape modeling performance is invariant under affine body shape transformations. Namely, in \cite{bib7} the invariant body shape modeling is formulated by finding the eigenvectors of the shape matrix of the body contour vectors and in \cite{bib6} the energy function minimized along the deformation of the contour model is properly selected so that the performed minimization is translation and scale invariant. But, in these approaches, shape transformation invariance concerns the performed energy minimization along contour-model deformation. Hence, in order to use such transformation invariant active contour models to obtain undeformed body contours given their deformed shapes, we should have had in advance a reliable estimation of the undeformed body shape. Otherwise, the curve to which contour model converges and the initial deformed body shape, are by no means, necessarily, in a one-to-one correspondence. These facts call upon an alternative approach which is, for the first time, presented in this paper.

Hence, in the present paper, we have introduced the following approach to tackle the problem of constructing identifiable images of undeformed body from its deformed instances: We estimate elastic deformation invariants (curves and sections) of a body suffering an equivalent to 2D deformation from an image of it at an arbitrary deformation instance. Knowledge of these invariants allows for unwrapping/straightening the deformed body image, a fact that in turn permits application of pattern recognition techniques for the body automatic classification/identification. We have applied the introduced approach to an important and some times crucial veterinary problem, namely the automatic identification of domestic animal parasites, from their images obtained via microscope \cite{bib18}. These images represent the parasites in a state of serious deformation. But the applicability of the introduced methodology seems to be quite more general, since the assumptions about the mechano-elastic properties of the deformed bodies that have been stated and adopted, are plausible and quite standard. In order to verify this, we have applied the introduced methodology to the determination of the undeformed state of elastic fibers, protozoa cells and lips expressions.

\section{A Brief Description of the Introduced Approach and the Resulting Application}
\label{sec2}
One of the major motivations for the present work was the fact that, in many important applications, one must identify objects on the basis of their deformed images. Very frequently, this is a very difficult task; for example, in the considered case of identifying parasites from images of them in deformed state, obtained via microscope, not even the expert parasitologist can identify the parasite gender. Thus, the authors developed an original methodology to tackle this type of problems which includes the following steps:\\
1. A set of general assumptions concerning the elastic deformation of the considered body was adopted. A main assumption, frequently encountered in elasticity theory, is that there are curves and sections whose characteristics remain invariant in any deformation stage.\\
2. A framework and a set of related equations concerning the body deformation process, have been developed, on the basis of the aforementioned assumptions. Since, many deformation invariant characteristics are associated with curves, the equations governing the body deformation have been written in curvilinear coordinates.\\
3. The previous analysis gave rise to two different approaches for straightening/unwrapping the deformed body. In the first approach, a major invariant curve, namely the neutral line, is determined together with its cross sections. The fact that the size of both the neutral line and its cross sections remain the same, leads to the construction of the undeformed body shape.
In the second approach the equations governing the body deformation are written in such a manner, so as to correspond to a sequence of fundamental image morphological operations. Therefore, the image of the undeformed body is obtained by application of the inverse procedure to the arbitrary deformed body image.\\
4. The system that was developed on the basis of the above approaches was applied to microscopic images of parasites suffering high deformation. The obtained images of straightened versions of the same parasite captured in different deformation instances, were very close to each other, thus confirming the validity and efficiency of the introduced approach (Section \ref{sec7}) (Figures \ref{fig5A}, \ref{fig5B}). In addition, the expert parasiologist has been able to fully identify the parasite gender from the undeformed images generated by the method (e.g. Figures \ref{fig9A}, \ref{fig9B}).\\ 
5. Finally, the authors proceeded in an automated recognition of the obtained straightened parasite versions. For this reason, they have developed a dedicated partitional classification algorithm that employs the Euclidean distance of the contours of the unwrapped body images. The representative contour curve of each group of parasites has been dynamically estimated via minimization of the classification error on a Training Set. The method offered an overall success identification rate of more than 97.6\%.

\section{Analysis of the Body Elastic Deformation}
\label{sec3}
\subsection{Adopted Assumptions Concerning the Considered Object's Elastic Properties}
\label{sec3_1}
\begin{enumerate}
\item All object (e.g. parasite) parts are isotropic, homogeneous and continuous, with a good approximation at least.
\item The static equation of balance holds for the deformed element too (1st order Theory).
\item There exists a piecewise smooth curve of symmetry for the undeformed object, or a curve of minimum curvature. We will use for it the term "reference curve"�.
\item Object's straight line segments, the cross sections, which are initially perpendicular to its symmetry curve, remain straight and perpendicular to a proper corresponding line after the deformation, which is usually called neutral line. The neutral line is always coplanar with the reference curve and, hence, the whole  deformation can be studied in two dimensions.
\item The generated stresses and displacements along the object body are linearly related, namely the generalized Hooke's law holds.
\end{enumerate}

The aforementioned hypotheses allow us to study the elastic behavior of the object in two dimensions. Consequently, the information extracted from the deformed object images may be sufficient for this study; for unwrapping the objects and eventually for automatically classifying them.
\subsection{Useful Fundamental Mechano-elastic Quantities}
\label{sec3_2}
We now proceed to summarize some notions that are commonly used in most approaches of Elasticity Theory. These will later be used in the analysis on which the unwrapping of deformed bodies is based.  In fact, in elasticity theory one defines the stress tensor $\hat \sigma = \begin{bmatrix} \sigma_{xx} & \sigma_{xy} \\ \sigma_{yx} & \sigma_{yy}\end{bmatrix}$, which expresses forces per unit length in all directions;  in $\sigma_{xx}$, $\sigma_{xy}$, $\sigma_{yx}$, $\sigma_{yy}$ the first subscript denotes the direction to which the stress is vertical, while the second subscript denotes the vector component direction. Thus $\sigma_{yx}$ is the x-component of the force per unit length exerted on a differential element vertical to the y-direction. Similarly one defines the strain tensor $ \hat \varepsilon = \begin{bmatrix} \varepsilon_{xx} & \tau_{xy} \\ \tau_{yx} & \varepsilon_{yy}\end{bmatrix}$, which expresses the relative elongations along all directions and possible torsions.

The body constitutive equation relates the stress tensor $\hat \sigma$ with the strain tensor $\hat \varepsilon$. These two tensors can be related through any functional form. However, in many practical circumstances, this functional form can be considered to be linear, namely $\hat \sigma = E \hat \epsilon$, where $E$ is a constant matrix (generalized Hooke's law). 
\subsection{Body's 2D Deformation Equations}
\label{sec3_3}
We assume that the undeformed object has a reference curve $M : \vec \mu(s) = \left(x_{\mu}(s) , y_{\mu}(s) \right)$ parameterized via its length $s$. Then we define the unit vector $\hat l (s) = \frac{\dot{\vec \mu}}{\left\| \dot{\vec\mu} \right\|}$ tangent at an arbitrary point of $M$ at length $s$ and the unit vector $\hat n(s)$ normal to $\hat l(s)$. Next, the coordinates of the points of object's body will be expressed via their distance vector from the reference curve $M$ and the length of this curve. Namely for each point with position vector $x \hat i + y \hat j$ we express this point via its distance vector from $M$.
\begin{equation}
\vec r(s,\delta) = \vec \mu(s)+ (x-x_{\mu}(s))\hat i +(y-y_{\mu}(s))\hat j = \vec \mu(s) + \delta \hat n(s)
\end{equation}
Thus any differential displacement in the undeformed body is written $d \vec r(s,\delta) = (1 +\delta \kappa(s))ds \hat l(s) + d \delta \hat n(s)$, where the curvature $\kappa(s) \hat l(s)= \frac{d}{ds}\hat n(s)$. 
\begin{figure}[!t]
\centering
\includegraphics[width=2.5in]{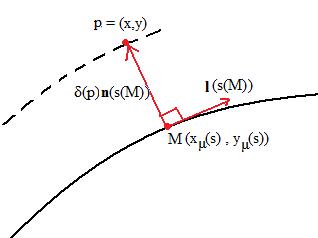}
\caption{Curvilinear coordinates $(s,\delta)$ of an arbitrary point p=(x,y) along a reference line.}
\end{figure}
Now, we can represent any differential deformation $du \hat i + dw \hat j$ by using the basis $(\hat l(s) , \hat n(s))$ as a curved vector.
\begin{equation}
du \hat i + dw \hat j = \left(1 + \delta_{u,w}\hat l(s)\frac{d \hat n(s)}{ds_{u,w}} \right)ds_{u,w}\hat l(s) + d \delta_{u,w} \hat n(s)
\end{equation}
\begin{equation}
ds_{u,w}=\sqrt{{\partial_s u}^2 + {\partial_s w}^2}=ds \sqrt{\left(\nabla^T u ~ \hat l(s)\right)^2 +\left(\nabla^T w ~ \hat l(s) \right)^2}
\end{equation}
\begin{equation}
d \delta_{u,w}=\sqrt{{\partial_{\delta} u}^2 + {\partial_{\delta} w}^2}=d \delta \sqrt{\left(\nabla^T u ~ \hat n(s)\right)^2 +\left(\nabla^T w ~ \hat n(s) \right)^2}
\end{equation}
Equivalently, one can define a displacement $(\partial_s \tilde u + \partial_s \tilde w) \hat l(s) + \partial_{\delta} \tilde w$, with the functions $\tilde u$ and $\tilde w$ corresponding to $u$ and $w$ via the expressions
\begin{eqnarray}
\nabla \tilde u &=& \hat l(s) \sqrt{\left(\nabla^T u ~ \hat l(s)\right)^2 +\left(\nabla^T w ~ \hat l(s) \right)^2} \nonumber \\
\nabla \tilde w &=& \hat n(s) \sqrt{\left(\nabla^T u ~ \hat n(s)\right)^2 +\left(\nabla^T w ~ \hat n(s) \right)^2}
\end{eqnarray}
Then the differential deformation is written 
\begin{eqnarray}
(\partial_s \tilde u + \partial_{s} \tilde w)\hat l(s) + \partial_{\delta} \tilde w \hat n(s)&=& \left(\left\| \nabla \tilde u \right\| + \tilde w \hat l^T(s)\frac{d \hat n(s)}{ds} \right)\nonumber \\ && ds \hat l(s) + \left\| \nabla \tilde w \right\| d \delta \hat n(s)
\label{3_3_1}
\end{eqnarray}
It also holds that
\begin{eqnarray}
\frac{d \hat n(s)}{ds}=\frac{d}{ds}\left(\frac{\nabla \tilde w}{\left\| \nabla \tilde w \right\|}\right)=\frac{H(\tilde w)-I \tilde w_{nn}}{\left\| \nabla \tilde w \right\|}\hat l(s)= \kappa(s) \hat l(s)
\label{3_3_2}
\end{eqnarray}
Then $\tilde u$ and $\tilde w$ are given by the differential equations
\begin{equation}
\frac{\partial \tilde w}{\partial \delta}=\left\| \nabla \tilde w \right\|
\label{3_3_3}
\end{equation}
\begin{equation}
\frac{\partial \tilde w}{\partial s}=- \tilde w \frac{\tilde w_{nn}}{\left\| \nabla \tilde w \right\|}
\label{3_3_4}
\end{equation}
\begin{equation}
\frac{\partial \tilde u}{\partial s}=\left\| \nabla \tilde u \right\|
\label{3_3_5}
\end{equation}
\begin{figure}[!t]
\centering
\includegraphics[width=2.5in]{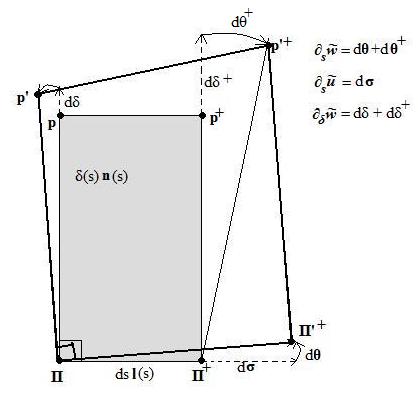}
\caption{Elementary deformation of a differential element $\Pi pp^{+}\Pi^{+}$. This element is placed on the undeformed reference  line at  length $s$. $\mathbf{l}$$(s)$ is the unit tangent vector of the symmetry line at this position and $\mathbf{n}$$(s)$ the unit normal one. The differential deformation of this element transforms it to $\Pi' p'p'^{+}\Pi '^{+}$. As mentioned above $\partial_s \tilde w$ describes differential rotation while $\partial_s \tilde u$ and $\partial_{\delta} \tilde w$ describe tangent and normal stretches respectively.}
\end{figure}
Strains $\varepsilon_{ll}$, $\varepsilon_{nn}$, $\varepsilon_{ln}$, $\varepsilon_{nl}$ caused by the deformation functions $\tilde u$, $\tilde w$ are given by the following expressions:
\begin{eqnarray}
\varepsilon_{ll}&=&\left\| \nabla \tilde u \right\|-1 \nonumber \\
\varepsilon_{nn}&=&\left\| \nabla \tilde w \right\|-1 \nonumber \\
\varepsilon_{ln}&=&\varepsilon_{nl}=-\frac{1}{2}(\tilde w - \delta_0)\frac{\tilde w_{nn}}{\left\| \nabla \tilde w \right\|} 
\end{eqnarray}
According to the Assumption 5, the relation connecting stresses and strains along $\hat l(s)$, $\hat n(s)$ is expressed via the relation $\tilde \sigma = \tilde E \tilde \varepsilon$; where $\tilde E = \begin{bmatrix} E_{ll} & E_{ln} \\ E_{nl} & E_{nn} \end{bmatrix}$ a constant matrix.
\subsection{Solution of Deformation Functions' PDEs Via Morphological Operations}
\label{sec3_4}
For an arbitrary point $(x,y)$ we consider the definitions\\
A)of its initial distance from a point $\Sigma$ of the reference line suffering s-deformation obtained via
\begin{equation}
\left\{\delta_0(x,y) \right\}(s)=\underset{\tau}{\min}\left\{\sqrt{\left(x-x_{\mu}(\tau) \right)^2+\left(y-y_{\mu}(\tau) \right)^2} \right\}
\label{3_4_1}
\end{equation}
B)of the correspondence of $\Sigma$ with a unique $\Sigma '$ of the reference line suffering $\delta$-deformation obtained via
\begin{equation}
\left\{s_0(x,y) \right\}(\delta)=\underset{\tau}{\arg \min}\left\{\sqrt{\left(x-x_{\mu}(\tau) \right)^2+\left(y-y_{\mu}(\tau) \right)^2} \right\}
\label{3_4_2}
\end{equation}
Then PDE (\ref{3_3_3}) with initial condition $\left\{\delta_0(x,y) \right\}(s)$ is equivalent to the action over $\left\{\delta_0(x,y) \right\}(s)$ of a dilation filter at scale $\delta$ with a flat disk kernel \cite{bib3}. Namely,
\begin{equation}
\tilde w(s,\delta) = M^{+}_{\delta}\left[\left\{\delta_0(x,y) \right\}(s)\right] = \underset{\left\|\vec \epsilon \right\| \leq \delta}{\sup}\left\{\left\{\delta_0 \left((x,y)+\vec \epsilon \right) \right\}(s) \right\}
\label{3_4_3}
\end{equation}
Similarly PDE (\ref{3_3_5}) with initial condition $\left\{s_0(x,y) \right\}(\delta)$ is equivalent to the dilation of $\left\{s_0(x,y) \right\}(\delta)$ with flat disk kernel at scale $s$.
\begin{equation}
\tilde u(s,\delta) = M^{+}_{s}\left[\left\{s_0(x,y) \right\}(\delta)\right] = \underset{\left\|\vec \epsilon \right\| \leq s}{\sup}\left\{\left\{s_0 \left((x,y)+\vec \epsilon \right) \right\}(\delta) \right\}
\label{3_4_4}
\end{equation}
Finally, equation (\ref{3_3_4}) can be written as $\frac{\partial}{\partial s}\ln{\tilde w} = - \frac{\partial}{\partial n}\left(\ln{\left\| \nabla \tilde w \right\|} \right)$, with initial conditions $\left\{\tilde w(x,y) \right\}(0,\delta)=M^{+}_{\delta}\left[\delta_0(x,y) \right]$ and $\left\{\tilde \phi_0(x,y) \right\}(0,\delta)=\ln \left(\left\| \nabla \tilde w  \right\| \right)(0,\delta)$. In Appendix A is shown that the above PDE problem is equivalent to constructing the image
\begin{equation}
\left\{\tilde w(x,y) \right\}(s,\delta)=\tilde w(0,\delta) \exp \left(\alpha \left[\tilde \phi_0(0,\delta)\right](s) \right)
\label{3_4_5}
\end{equation}
Where the filter $\alpha \left[g(\delta) \right](s)$ is defined by the formula
\begin{equation}
\alpha \left[g(\delta) \right](s)=\sup \left[M^{-}_s[g(\delta)] , \inf \left[M^{+}_{s}[g(\delta)] , \kappa_{\phi}(s_0(\delta)) \right] \right]
\label{3_4_6}
\end{equation}
where $M^{-}_s[g(\delta)] = \underset{\left\|\vec \epsilon \right\| \leq s}{\inf}\left\{g((x,y)+ \vec \epsilon)(\delta) \right\}$ and the reference image $\left\{\kappa_{\phi}(x,y) \right\}(s_0(\delta)) = - \mbox{sgn} \left( \kappa(s_0(\delta)) \right)\infty$. The $- \infty$ normalization is used in order to obtain the identical element of $\inf$
\subsection{How the previous results lead to obtaining undeformed body images}
\label{sec3_5}
We have reformulated some fundamental relations that describe the 2D body deformation, in order to derive equations for the deformation process which can be subsequently interpreted as a set of morphological operations acting on a 2D body image. 

Suppose that the unknown image of the undeformed 2D body is $I_U$; evidently, $I_U$ is unavailable. On the contrary, the only available information is an image of the deformed body, say $I_D$. We have proved that the deformation of the body satisfies equations (\ref{3_3_3}), (\ref{3_3_4}), (\ref{3_3_5}). This is equivalent to having proved that $I_D$ can be obtained from $I_U$ by application of morphological operations on $I_U$ described in (\ref{3_4_3}), (\ref{3_4_4}), (\ref{3_4_5}).

Since only $I_D$ is available, by applying the inverse morphological operation we obtain the undeformed image of the body via the process below:\\
Initially we create images $\delta_0= \left\{\delta_0(x,y) \right\}(s_0)=\underset{\tau}{\min}\left\{\sqrt{(x-x_{\mu}(\tau))^2 + (y-y_{\mu}(\tau))^2} \right\}$, $s_0= \left\{s_0(x,y) \right\}(\delta_0)=\underset{\tau}{\arg \min}\left\{(x-x_{\mu}(\tau))^2 + (y-y_{\mu}(\tau))^2 \right\}$ which will play the role of initial conditions for the subsequent process:\\ A) First we perform dilation of image $s_0$ until the adopted $s$-scale of deformation and we produce the image $\left\{\tilde s(x,y) \right\}(s) = \tilde u[s_0](s)=M^{+}_s(s_0)$. Successively, we perform dilation of image $\delta_0$ until the adopted $\delta$-scale of deformation and we produce the image
\begin{equation}
\left\{\tilde \delta(x,y) \right\}(\delta) = M^{+}_{\delta}[\delta_0]
\label{3_4_7}
\end{equation}
B)Next we use this image of $\delta$-deformations to produce image
\begin{equation}
\left\{\tilde \phi_0(x,y) \right\}(\delta) = \ln \left\| \nabla \tilde \delta(\delta) \right\| = \ln \left(\frac{\partial}{\partial \delta}\tilde \delta(\delta) \right) 
\label{3_4_8}
\end{equation}
on which, subsequently, the following operation is performed.\\
C) We consider that the image $\left\{\tilde w(x,y) \right\}(s,\delta)$ is a product of two independent operations $\left\{\tilde \phi(x,y) \right\}(s)$, $\left\{\tilde \delta(x,y) \right\}(\delta)$. Where $\left\{\tilde \delta(x,y) \right\}(\delta)$ is the deformation that $\left\{\tilde w(x,y) \right\}(s,\delta)$ performs along $\hat n(s)$ and it can be immediately obtained from A); while $\left\{\tilde \phi(x,y) \right\}(s)$ is the deformation of the body because of the change of reference line's curvature. Then, we determine the image $\left\{\tilde \phi(x,y) \right\}(s)$ from the solution of $\left\{\tilde w(x,y) \right\}(s,\delta)$. Namely, $\left\{\tilde w(x,y) \right\}(s,\delta) = \left\{\tilde \delta(x,y) \right\}(\delta) \exp \left(\alpha[g(\delta)](s) \right)$ results
\begin{equation}
\left\{\tilde \phi(x,y) \right\}(s) =\exp \left(\alpha[g(\delta)](s) \right) 
\label{3_4_9}
\end{equation}
where $g(\delta)=\left\{\tilde \phi_0(x,y) \right\}(\delta)$ has been obtained by (\ref{3_4_8}).

Next, on the basis on Section \ref{sec3} analysis, two approaches are analytically described in order to construct unwrapped body versions from its deformed images. 

\section{The First Approach - Determination Of Deformed Body's Neutral Line}
\label{sec4}
In the present section we will introduce a novel method for the determination of the neutral line in the deformed body; we once more emphasize that along the neutral line no stress is exerted and that this curve corresponds to the reference line of the undeformed body, as we will demonstrate below.

\subsection{Confirmation that the neutral line passes from the middle of its cross sections}
\label{sec4_1}
Consider a section $A$ normal to the reference line in the undeformed object's state. Then, the vertical shear force $V$ and force $N$ normal to $A$ and along the reference line are given by the expressions: $V=\oint_A\sigma_{ln}dA$ and $V=\oint_A\sigma_{ll}dA$

We assume that the body is in an equilibrium position in each image of its deformation instances. Then, equilibrium along reference line implies that $N=0$, $V=0$. Since we have adopted the assumption that the generalized Hooke's law holds, it follows that 
\begin{eqnarray}
N&=&E_{ll}\left(\left\| \nabla \tilde u \right\|-1 \right)A -\frac{E_{ln}}{2}\frac{\tilde w_{nn}}{\left\| \nabla \tilde w \right\|}\oint_A(\tilde w -\delta_0)dA \nonumber \\ &=&E_{ll}\left(\left\| \nabla \tilde u \right\|-1 \right)A
\label{4_1_3}
\end{eqnarray}
Combining (\ref{4_1_3}) and $N=0$, we deduce that 
\begin{equation}
\left\{\tilde s(x,y) \right\}(\delta)=s_0+s
\label{4_1_4}
\end{equation}
indicating that $s$-deformation is only offsetting the length of the initial reference line and hence that it can be ignored. Now, reference line's stresses in the direction tangent to it, $\sigma_{ll}$, are given by 
\begin{equation}
\sigma_{ll}=-\frac{E_{ln}}{2}(\tilde w -\delta_0) \frac{\tilde w_{nn}}{\left\| \nabla \tilde w \right\|}
\label{4_1_5}
\end{equation}
\begin{eqnarray}
V&=&E_{ln}\oint_A \left(\left\| \nabla \tilde w \right\|-1 \right)dA -\frac{E_{ll}}{2}\frac{\tilde w_{nn}}{\left\| \nabla \tilde w \right\|}\oint_A(\tilde w -\delta_0)\nonumber \\ &=&E_{ln}\oint_A \left(\left\| \nabla \tilde w \right\|-1 \right)dA
\label{4_1_6}
\end{eqnarray}
Combining (\ref{4_1_5}) and $V=0$, we deduce that
\begin{equation}
\left\{\tilde \delta(x,y) \right\}(\delta)=\delta_0+\delta
\label{4_1_7}
\end{equation}
which similarly  results that $\delta$-deformation is only offsetting the distances from the undeformed reference line and hence that it can be ignored. Now, reference line's stresses in the directions normal to it, $\sigma_{ln}$, are given via 
\begin{equation}
\sigma_{ln}=-\frac{E_{ll}}{2}\frac{\tilde w_{nn}}{\left\| \nabla \tilde w \right\|}(\tilde w - \delta_0)
\label{4_1_8}
\end{equation}
Combining (\ref{4_1_8}) and (\ref{4_1_5}) with zero offsetting of $\delta_0(x,y)$ we obtain $\sigma_{ll}=-\frac{E_{ln}}{2}\left\{\tilde \phi(x,y) \right\}(s) \delta_0 \frac{\tilde w_{nn}}{\left\| \nabla \tilde w \right\|}$, $\sigma_{ln}=-\frac{E_{ll}}{2}\left\{\tilde \phi(x,y) \right\}(s) \delta_0 \frac{\tilde w_{nn}}{\left\| \nabla \tilde w \right\|}$.

To study the stress that reference line suffers we let $\delta_0 \to 0$. Then (\ref{3_4_8}) implies $\left\{\tilde \phi_0(x,y) \right\}(\delta) = \underset{\delta_0 \to 0}{\lim}\ln \left(\frac{\partial}{\partial n}\delta_0(x,y) \right)=0 \Rightarrow \underset{\delta_0 \to 0}{\lim}\left\{\tilde \phi(x,y) \right\}(s)=1$ and finally (\ref{3_4_9}) results in $\underset{\delta_0 \to 0}{\lim}\left\{\tilde w(x,y) \right\}(s,\delta)=\underset{\delta_0 \to 0}{\lim}\delta_0 \left\{\tilde \phi(x,y) \right\}(s)=0$.
Therefore, the curve to which symmetry line is transformed suffers no stress; the curve formed by all these unstressed points, usually called the neutral line, has the following properties:\\
A) It is the curve to which the reference line is transformed due to the elastic deformation process.\\
B) No stress is exerted along it.\\
C) As a consequence, the neutral line and the body reference line are of the same length.\\
D) The neutral line passes from the middle point of each cross section in the 2D image representation of the deformed body.\\
E) The undeformed cross sections initially perpendicular to the reference line, remain perpendicular to the neutral line even after the body deformation.
\subsection{An introduced important property of the deformed object contour tangents}
\label{sec4_2}
In this paragraph we will state a new lemma that will play an important role for the determination of the neutral line and the unwrapping of the body. The proof of this lemma is given in Appendix B.\\
\textit{Lemmma}\\
Let $\Sigma_1 \Sigma_2$ be the symmetry line of the unwrapped body and $AD$ an arbitrary cross section, intersecting $\Sigma_1 \Sigma_2$ at point $M$, where part $\Sigma_1 M$ of the symmetry line has length $s$. Then, due to the deformation, $AD$ moves to a section $A'D'$, perpendicular to the neutral line at point $M'$. Let moreover $U’$ and $L’$ be the upper and lower boundary curves of the deformed body, respectively and $M'N'$ be its neutral line differential element. If $\vec T_{A'}^{U'}$ is the tangent vector of $U'$ at $A'$, $\vec T_{D'}^{L'}$ the tangent vector of $L'$ at $D'$ and $\vec L_{M'}$ is the tangent vector of the deformed symmetry line at point $M'$; then it holds that $\angle \left(\vec{A'D'} , \vec T^{U'}_{A'}-\vec L_{M'} \right) + \angle \left(\vec{A'D'} , \vec T^{L'}_{D'}-\vec L_{M'} \right) = \pi$.

This lemma proves to be fundamental in defining and obtaining the cross section's coordinates, as well as those of the neutral line. Knowledge of the coordinates of the neutral line allows us to "unwrap" the outline of the deformed body, as is shown in Sects. \ref{sec4_4}, \ref{sec4_5}.
\begin{figure*}[!t]
\centering
\includegraphics[width=5in]{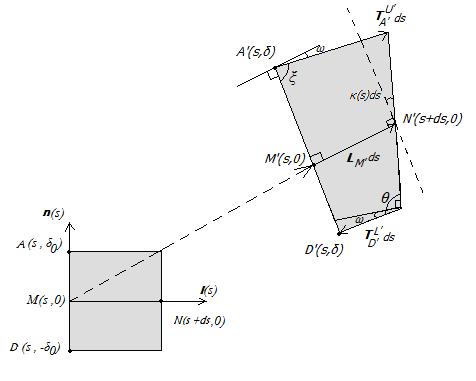}
\caption{Interpretation of the Lemma. The arbitrary cross section $AD$ in the undeformed parasite body moves to $A'D'$ via the deformation process. $U'$ denotes the body upper contour and $L'$ the lower contour. $\mathbf{T}_{A'}^{U'}$ is the tangent vector of $U'$ at $A'$, $\mathbf{T}_{D'}^{L'}$ the tangent vector of $L'$ at $D'$ and $\mathbf{L}_{M'}$ is the tangent vector of the deformed symmetry line at point $M'$ of the deformed neutral line. We determine body differential element's cross section via the demand that $\xi + \theta = \pi$. Notice that in a neighborhood of $D'$ there is no other point satisfying $\xi + \theta = \pi$ since $\theta$ remains fixed and $\xi$ changes.}
\end{figure*}
\subsection{Object Contour Extraction And Its Polynomial Approximation}
\label{sec4_3}
As the introduced methodology makes use of the border line of each parasite, it is necessary to obtain well-defined boundary lines of all instances of the examined parasites. In order to achieve this, the following method is used:

First, we have applied various image segmentation methods (\cite{bib1}, \cite{bib8}) in order to obtain quite clear-cut and accurate region borders of each parasite. A rather simple method that seems to work well, for the considered images, is the one that uses each parasite's pixel intensity histogram and the lower turning point of it. All pixels with intensity lower than this turning point may be considered to belong to the parasite body. This method may generate various artifacts that may be removed by application of proper morphological filters, see e.g. \cite{bib12}. 

As it will become evident from the subsequent analysis, in order that the introduced methodology is applied, each contour line must have the following properties: a) each pixel must have exactly two neighboring pixels, b) no isolated pixels or groups of pixels are allowed and c) three pixels must not form a compact right ($90^o$) angle. Since no edge detection algorithm can generate the parasite contour in this form, suitable software has been developed to achieve that.
The next step is to determine if there are specific mathematical curves that optimally fit the object body, e.g. a parasite, contour in the obtained images. There are various techniques for achieving this goal (e.g. \cite{bib2}, \cite{bib5}, \cite{bib13}). For the present application the following method proved quiet satisfactory: 

	The curve parameter is chosen to be its contour length s, calculated via the distance of the successive pixels that form it. Subsequently, we approximate the variables x and y of the body contour by polynomials of the appropriate degree, estimated by means of the body contour curvature: 
\begin{equation}
x(s) = \sum_{n=0}^N a_n s^n~,\; y(s) = \sum_{n=0}^N b_n s^n
\label{4_3_1}
\end{equation}
Suppose now that one wants to test if the upper body contour corresponds to the aforementioned curve and at the same time to compute the parameters of this optimal curve: 
Let $\vec r^B_i$, $i=1,2,...,N^B$ be the centers of the pixels forming the upper contour and let $\Pi=\left[a_n,...,a_0,b_n,...,b_0 \right]$. Hence the parametric vector equation of the prototype curve is $\vec r^M(s|\Pi) = x(s) \hat i + y(s) \hat j$.

Next, we compute the optimal set of parameters $\Pi^O$ and the corresponding sequence of values of the independent variable $s_i$, $i=1,2,...,N^B$, so that $\vec r^M(s_i|\Pi^O)$ best fits $\vec r^B_i$ according to the chosen norm 
$E_2=\sum_{i=1}^{N^B}\left\| \vec r^B_i - \vec r^M_i \right\|$.

\subsection{The Developed Algorithm for Determining the Deformed Body's Neutral Line}
\label{sec4_4}
In this section, we will analytically describe the methodology we have introduced and applied for determining the exact position of the neutral line in the deformed body image, as well as the positions of the cross sections that, by hypothesis, always remain undeformed and normal to the neutral line. The application of this methodology has been made on the available parasite images and comprises the following steps:\\
\textit{\underline{Step 1}} - We, first, extract the parasite contour (Sect. \ref{sec4_3}). Next, we spot the head and the tail of the parasite as follows: First, to spot the tail, for each contour pixel $p$ we consider the sets of pixels $P_L$ that lie on its left and $P_R$ that lie on its right. We approximate both $P_L$ and $P_R$ with line segments in the Least Squares sense. We let the tail $T$ be the pixel where these two line segments form the most acute angle.

Second, we spot the parasite "head" $H$: We move away from the tail and we locally approximate the contour by polynomials of fifth degree, of which we compute the curvature. We let the "head" be the point of maximum curvature which also lies between 0.4 and 0.6 of the whole contour length. \\
\textit{\underline{Step 2}} - We divide the whole contour into two parts I and II (arbitrarily upper and lower), that both end at the parasite "head" and "tail"�. Then we approximate both parts with polynomials of type (\ref{4_3_1}). All performed experiments indicate that this approximation is excellent. We form a dense sequence $M^{II}_j$, $j=1...N^{II}$ of points belonging to the polynomial curve best fitting part II and a less dense sequence $M^I_i$, $i=1...N^I$ on the curve fitting part I; let $\vec \tau^I_i$ and $\vec \tau^{II}_j$ be the unit tangent vectors to these model curves at each $M^I_i$ and $M^{II}_j$ respectively.\\
\textit{\underline{Step 3}} - Subsequently, we spot parasite's neutral line by applying the aforementioned lemma as follows: We move away from tail $T$ along part I and we connect $M^I_1$ with each point of set $M^{II}_j$, $j=1...K$, where $K$ is a predefined number of pixels, say 5\% of the whole contour length. We form vectors $\vec r_{1,j}= \vec{M^{II}_j M^I_1}$. We keep only those vectors $\vec r_{1,j}$ that lie entirely within the parasite body and for these we compute the angles $\varphi^I_{1,j}$, $\varphi^{II}_{1,j}$ formed by each vector $\vec r_{1,j}$ and the tangent vectors $\vec \tau^I_i$ and $\vec \tau^{II}_{1,j}$ respectively. Then, we define the sequence $\Delta \varphi_{1,j}= \left|\varphi^I_{1,j}+\varphi^{II}_{1,j}-\pi \right|$ and we let $N^{II}_1$ be that point where the minimum value of the sequence $\Delta \varphi_{1,j}$ occurs, say the $d_1$-th of the sequence $M^{II}_j$; we consequently define $\vec{M^I_1 N^{II}_1}$ to be a cross section of the parasite that remains undeformed and normal to the neutral line.

Next we compute the second cross section as follows: We move away from the tail vertex $T$, and $M^I_1$, at $M^I_2$ and once more, we define the set of points $M^{II}_{d_1+j}$, $j=1...K$. Proceeding as before we define vectors $\vec r_{2,j}= \vec{M^{II}_{d_1+j}M^I_2}$, $j=1,...,K$. We compute the corresponding angles $\varphi^I_{2,j}$ between $\vec r_{2,j}$ and $\vec \tau^I_2$, as well as $\varphi^{II}_{2,j}$ between $\vec r_{2,j}$ and $\vec \tau^{II}_{d_1+j}$. We spot the minimum of the sequence $\Delta \varphi_{2,j}= \left|\varphi^I_{2,j}+\varphi^{II}_{2,j}-\pi \right|$ which occurs at point $N^{II}_2$, say the $d_2$-th point of sequence $M^{II}_j$. We let $\vec{M^I_2 N^{II}_2}$ be the second cross section that remains undeformed and normal to the neutral line.

\begin{figure}[!t]
\centering
\includegraphics[width=2.5in]{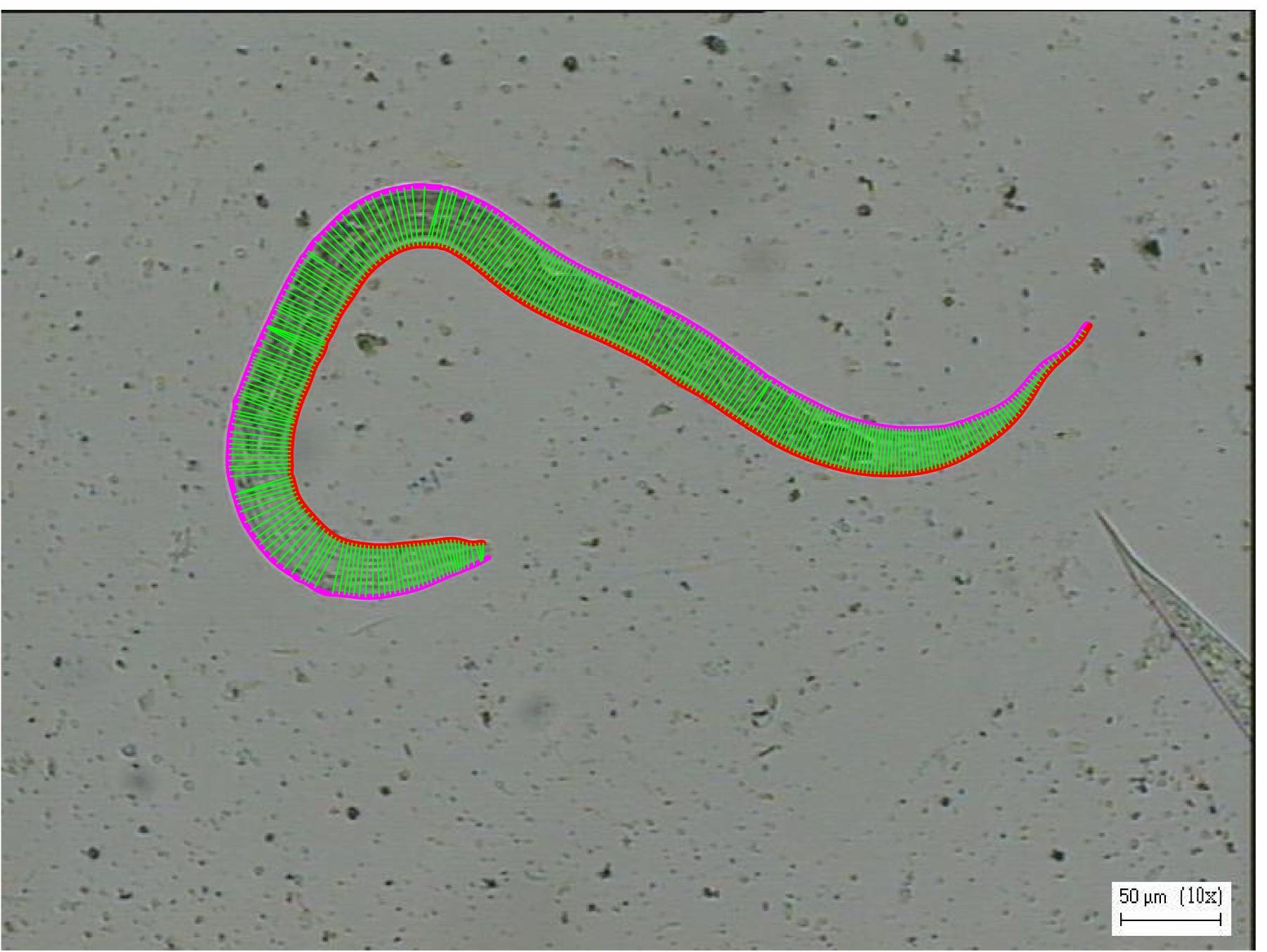}
\caption{Determination of the cross sections in a highly deformed parasite image. A first example.}
\label{fig4A}
\end{figure}

\begin{figure}[!t]
\centering
\includegraphics[width=2.5in]{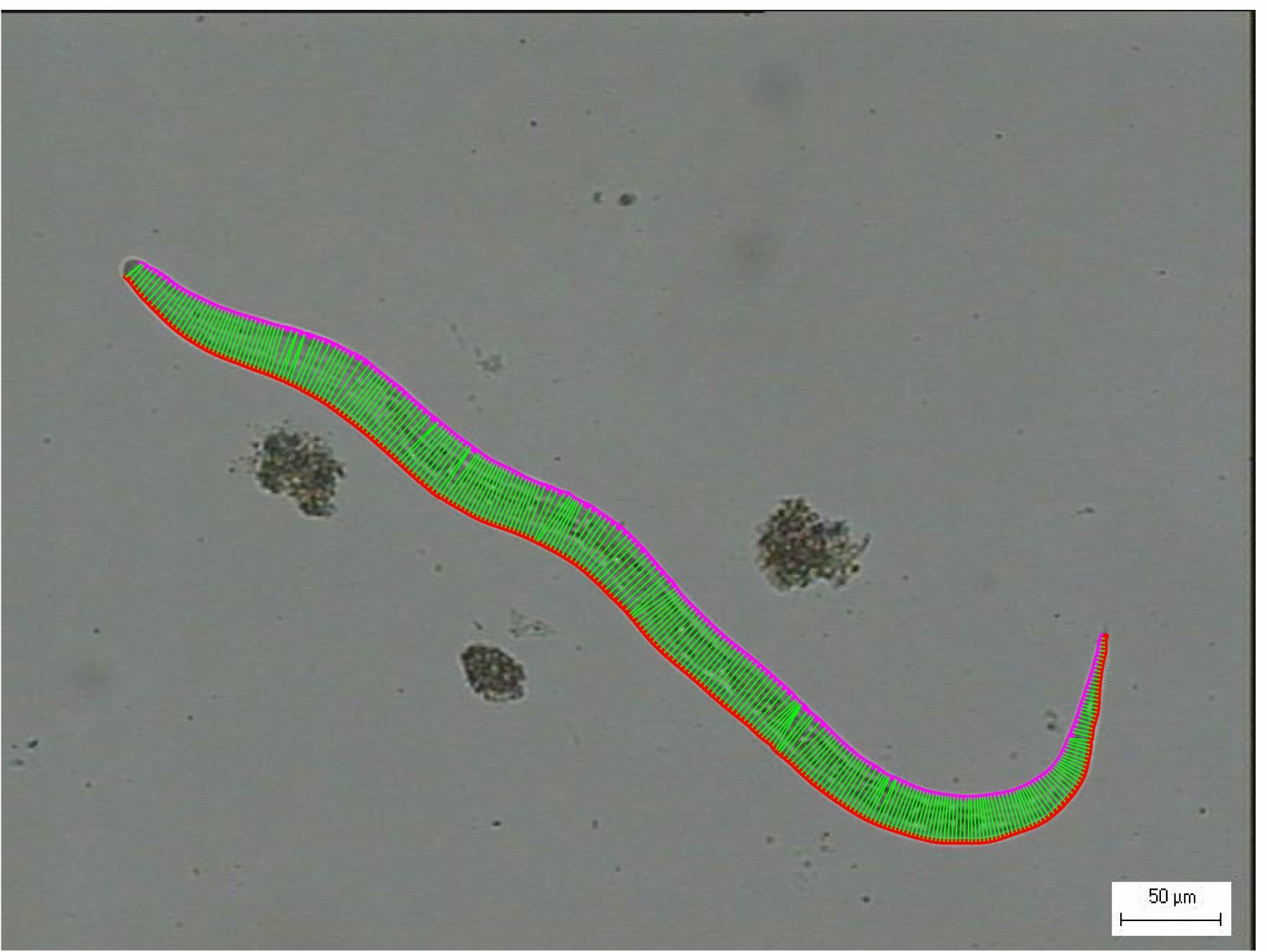}
\caption{Another example of cross sections determination}
\label{fig4B}
\end{figure}

Finally we proceed in obtaining all cross sections $\vec{M^I_i N^{II}_i}$ passing from $M^I_i$, $i=1...N^I$ by the same method (Figures \ref{fig4A}, \ref{fig4B}). The middle points of these cross sections belong to the neutral line and the unit vector normal to these sections is tangent to the neutral line. 
\subsection{Unwrapping The Deformed Object}
\label{sec4_5}
We have shown in Section \ref{sec4_1} that under the adopted assumptions the neutral line undertakes no stress and it is found in the middle of the corresponding  cross sections. Consequently, we define the neutral line of the deformed parasite to be the locus of the middle points $K_{\nu}$ of the cross sections $M^I_{\nu}N^{II}_{\nu}$ as they are determined above. Clearly, the length of the neutral line remains unchanged during the various phases of the body deformation and it coincides with the length of the undeformed body's symmetry axis.

\begin{figure}[!t]
\centering
\includegraphics[width=2.5in]{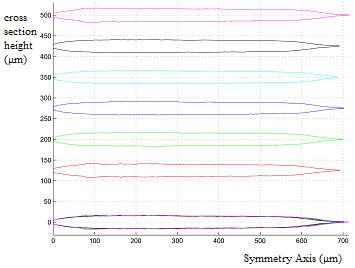}
\caption{ Contours of the unwrapped versions obtained from six
different instances of larva deformations of the same individual}
\label{fig5A}
\end{figure}

\begin{figure}[!t]
\centering
\includegraphics[width=2.5in]{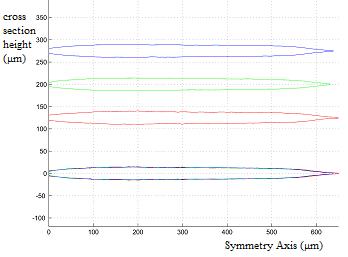}
\caption{Contours  of  the  unwrapped  versions obtained  from three different instances of another individual.}
\label{fig5B}
\end{figure}

 Therefore, in order to straighten the body and find its undeformed shape, we proceed as follows: 
First, we compute the distance of all successive middle points $K_i$ and $K_{i+1}$. Then, along the $x$-axis we form a sequence of points $\Lambda_i$ of equal number with $K_i$ as follows: $\Lambda_1$ is placed in the axis origin; $\Lambda_2$ in the positive $x$-axis, so that the distance between points $\Lambda_1$ and $\Lambda_2$ is equal to the distance between middle points $K_1$ and $K_2$. We continue this process so that $\Lambda_i$ and $\Lambda_{i+1}$ are equidistant with $K_i K_{i+1}$, until all middle points are exhausted. Moving in a direction perpendicular to the $x$-axis at each point $\Lambda_i$ we choose two points: $A_i$ with $y$-coordinate equal to $\lambda_i/2$, namely half the length of the cross section $M_i^I N_i^{II}$ and point $B_i$ with $y$-coordinate $-\lambda_i/2$. We set points $A_i$ to form the one part of the parasite,
while points $B_i$ form the other parasite part (Figures \ref{fig5A}, \ref{fig5B}).
\section{The Second Approach - Determining The Undeformed Body Shape Via Filtering Operations On Its Deformed Image}
\label{sec5}
The method introduced in this section utilizes the solution of 2D body deformation PDEs as it was formulated in Section \ref{sec3_5}, in order to construct the inverse deformation for the considered deformed body image. Also, exploitation of Assumption 2, in Section \ref{sec4_1}, has simplified this solution, since there, it has been shown that solution's Step A can be ignored. So, the sequence of image operations that construct the unwrapped body version is described in the following subsection.
\subsection{Description Of The Proposed Method. Application To The Unwrapping Of Deformed Parasites.}
\label{sec5_1}
\textit{\underline{Step 1}}: Initially, the vector parametric equation of the deformed body contour is approximated via polynomials of type (\ref{4_3_1}) in the least squares sense, thus obtaining $\vec r_c(s) = \left(x_c(s) , y_c(s) \right)$. Using this vector parametric equation, we compute the unit directional vectors along $\vec r_c(s)$, $\hat l(s) = \frac{\left(\dot x_c(s) , \dot y_c(s)\right)}{\sqrt{\dot x_c(s)^2 + \dot y_c(s)^2}}$, $\hat n(s) = \frac{\left(-\dot y_c(s) , \dot x_c(s)\right)}{\sqrt{\dot x_c(s)^2 + \dot y_c(s)^2}}$, as well as the curvature $c(s)=\frac{\dot x_c \ddot y_c - \dot y_c \ddot x_c}{\left(\dot x_c^2 + \dot y_c^2 \right)^{3/2}}$.\\
\textit{\underline{Step 2}}: Afterwards, at any pixel point $(x,y)$, inside the deformed body, we attribute the values the following two images, $s_0(x,y)$, $\delta_0(x,y)$, have at this point $\delta_0(x,y) = \underset{s}{\min}\left\{\sqrt{(x-x_c(s))^2+(y-y_c(s))^2} \right\}$, $s_0(x,y) = \underset{s}{\arg \min}\left\{\sqrt{(x-x_c(s))^2+(y-y_c(s))^2} \right\}$.  These two images, $s_0(x,y)$, $\delta_0(x,y)$, play the role of initial conditions for the PDEs (\ref{3_3_3}), (\ref{3_3_5}), (\ref{3_3_4}) which describe the deformation functionals $\left\{\tilde \delta(x,y) \right\}(\delta)$, $\left\{\tilde s(x,y) \right\}(s)$, $\left\{\tilde \phi(x,y) \right\}(s)$ respectively.
\begin{figure}[!t]
\centering
\includegraphics[width=2.5in]{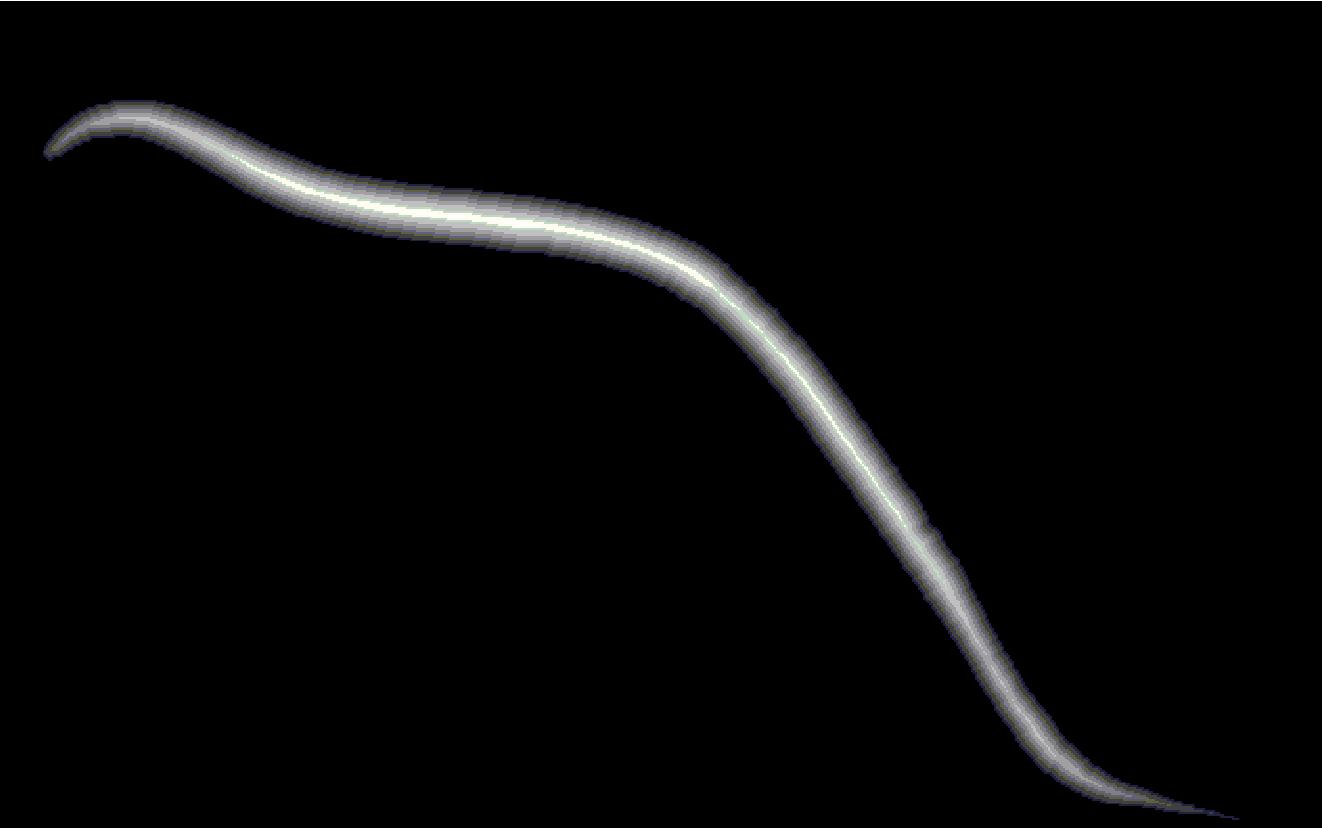}
\caption{Image, where the parasite body pixels have intensity $\delta_0(x,y)$}
\label{fig6}
\end{figure}
\\
\textit{\underline{Step 3}}: We have proved in Section \ref{sec3_4} that the solution of these deformation differential equations is equivalent to applying dilations (\ref{3_3_5})
and (\ref{3_3_3}) to $s_0(x,y)$, $\delta_0(x,y)$ respectively and the filter $a[g , \kappa](s)$ to the dilated $\delta_0(x,y)$ (see Figure \ref{fig6}) as formula (\ref{3_3_4}) indicates. As it was shown in Section \ref{sec4_1}, the crucial operation is curvature deformation $\left\{\tilde \phi(x,y) \right\}(s)$ given by formula (\ref{3_4_9}). Namely $\left\{\tilde \phi(x,y) \right\}(s) = \exp \left(\alpha \left[g , \kappa_{\phi} \right](s) \right)$, where $\kappa_{\phi}(x,y) = - \mbox{sgn}\left(c(s_0(x,y)) \right) \infty$, $g(x,y) = \ln \tilde \phi_0(x,y)$ and $\tilde \phi_0(x,y)=\left\| \nabla \delta_0(x,y) \right\|$. 
\begin{figure}[!t]
\centering
\includegraphics[width=2.5in]{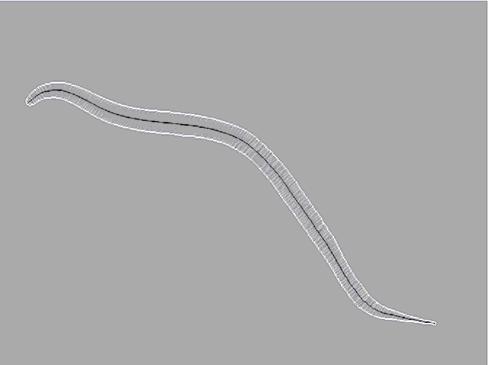}
\caption{Image of parasite body curvature deformation minimized at a scale $\sigma(x,y)$}
\label{fig7}
\end{figure}
This image (Fig. \ref{fig7}) represents all curvature deformations applied on the body, through scale $s$. Since we want to straighten parasite bodies we demand zero curvature for the reference line. Thus, for each pixel $(x,y)$ in the deformed parasite body, we determine the proper scale $\sigma(x,y)$ that minimizes $\left\{\phi(x,y) \right\}(\sigma)$, namely $\sigma(x,y) = \underset{s}{\arg \min}\left\{\left\{\phi(x,y) \right\}(s) \right\}$ until a curve of a threshold curvature deformation is spotted (see Figures \ref{fig7}, \ref{fig8}). 
\begin{figure}[!t]
\centering
\includegraphics[width=2.5in]{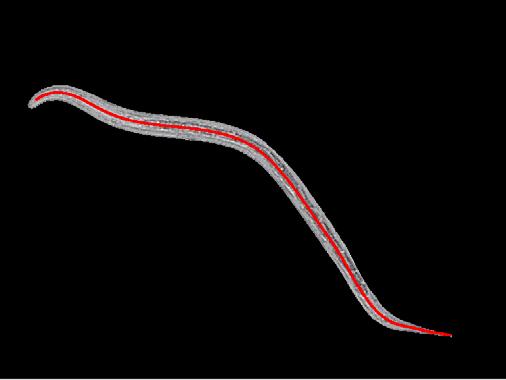}
\caption{Parasite body reference curve of approximately zero curvature deformation estimated after inversion.}
\label{fig8}
\end{figure}
Consider that this curve consists of points $(x_{\mu} , y_{\mu})$. Finally, if the given image of the deformed body is $f(x,y)$, then the above procedure generates an image $f_T(\tilde x , \tilde y)=f(x,y)$, $\tilde x = M^{+}_{\sigma(x,y)}[s_0(x,y)]$, $\tilde y = \delta_0(x,y) - \delta_0(x_{\mu},y_{\mu})$ (Figures \ref{fig9A}, \ref{fig9B}).
\begin{figure}[!t]
\centering
\includegraphics[width=2.5in]{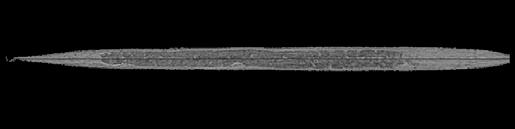}
\caption{Generating the unwrapped version of the parasite of Fig. \ref{fig4A}.}
\label{fig9A}
\end{figure}
\begin{figure}[!t]
\centering
\includegraphics[width=2.5in]{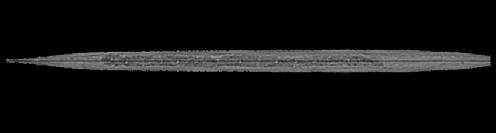}
\caption{Generating the unwrapped version of the parasite of Fig. \ref{fig4B}.}
\label{fig9B}
\end{figure}

\subsection{Application of the method to the determination of the undeformed shape of other bodies}
\label{sec5_2}
  We have applied the method described above to three different cases of body deformation, namely the deformation of cells (protozoa), in human lips deformation and in elastic fibers deformation.

\begin{figure}[!t]
\centerline{\subfloat[]{\includegraphics[width=1.5in]{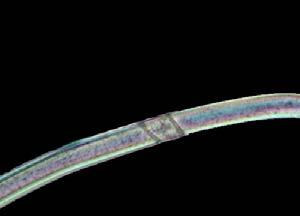}
\label{fig10a}}
\hfil
\subfloat[]{\includegraphics[width=1.5in]{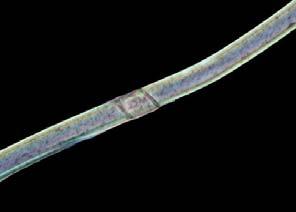}
\label{fig10b}}}
\vfil
\centerline{
\subfloat[]{\includegraphics[width=1.5in]{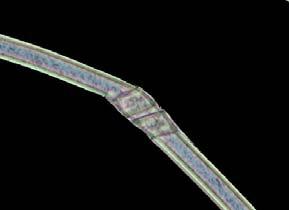}
\label{fig10c}}
}
\caption{Segmented microscopic images of three deformed elastic fibers of the same type and structure.}
\label{fig10}
\end{figure}

\begin{figure}[!t]
\centerline{\subfloat[]{\includegraphics[width=1.75in]{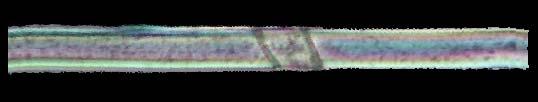}
\label{fig11a}}
\hfil
\subfloat[]{\includegraphics[width=1.75in]{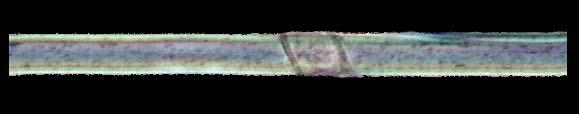}
\label{fig11b}}}
\vfil
\centerline{
\subfloat[]{\includegraphics[width=1.75in]{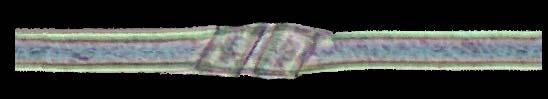}
\label{fig11c}}
}
\caption{Straightened versions of the fibers of Fig. \ref{fig10}, presented in the same order with the segmented images.}
\label{fig11}
\end{figure}

In fact, the case of elastic fiber deformation (fig. \ref{fig10}) can be tackled by immediate application of the method described in Sect. \ref{sec5_1} above, since the fibers have a symmetry axis in their undeformed state. We notice that both methods described in Sections \ref{sec4}, \ref{sec5_1} can be applied for the determination of elastic fibers undeformed shape (fig. \ref{fig11}).

The case of cell undeformed shape restoration can be treated by means of the method described in Sect. \ref{sec5_1}, since in the undeformed state the cells bear a reference curve, which is not necessarily a symmetry axis. In other words, Assumptions 1,2,4,5 stated in Section \ref{sec3_1} hold intact both in this and in the elastic fibers case; concerning Assumption 3, we note that in the case of the fibers the reference curve is also a zero curvature symmetry axis, while in the cells case this condition is relaxed. We note that for the restoration of the undeformed shape of the cells considered here, we have employed the exact method described in Sect. \ref{sec5_1} above, using elliptical coordinates for the description of the undeformed body points (Fig. \ref{fig13}).
\begin{figure*}[!t]
\centerline{
\subfloat[]{\includegraphics[width=1.5in]{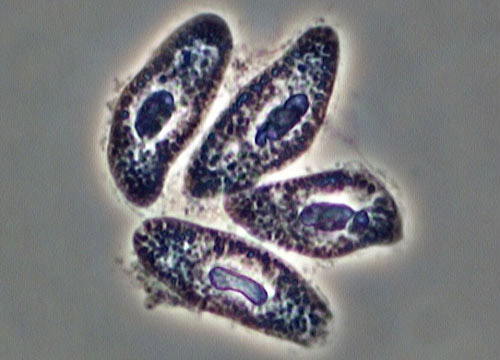}
\label{fig12or}}
\hfil
\subfloat[]{\includegraphics[width=1.5in]{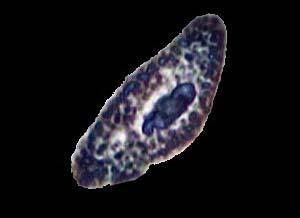}
\label{fig12a}}
\hfil
\subfloat[]{\includegraphics[width=1in]{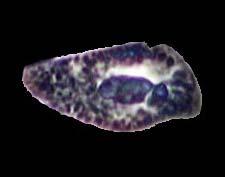}
\label{fig12b}}
\hfil
\subfloat[]{\includegraphics[width=1.5in]{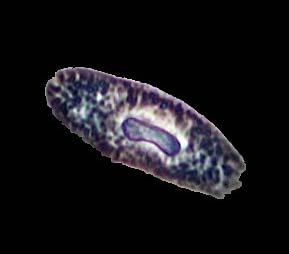}
\label{fig12c}}
\hfil
\subfloat[]{\includegraphics[width=1in]{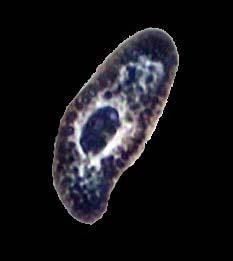}
\label{fig12d}}
}
\caption{The original image \ref{fig12or} and the segmented microscopic images of four individuals of the same protozoon in a deformed state \ref{fig12a}-\ref{fig12d}.}
\label{fig12}
\end{figure*}
\begin{figure}
\centering
\subfloat[]{\includegraphics[width=1.5in]{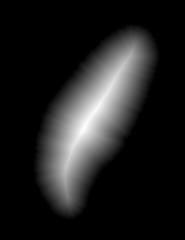}
\label{fig13a}}
\hfil
\subfloat[]{\includegraphics[width=1.5in]{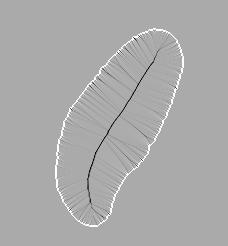}
\label{fig13b}}\\
\subfloat[]{\includegraphics[width=1.5in]{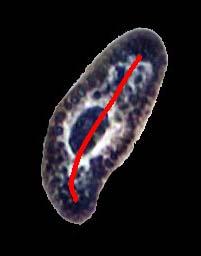}
\label{fig13c}}
\hfil
\subfloat[]{\includegraphics[width=1.25in]{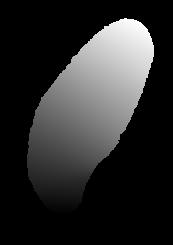}
\label{fig13d}}
\caption{Intermediate steps of the deformation inversion process of Sect. \ref{sec5_1} for the individual of Fig. \ref{fig12d}\newline
a)Initialization of the filtering process; image, where the cell body pixels have intensity $\delta_0(x,y)$.\newline
b)Image of cell body curvature deformation minimized at a scale $\sigma(x,y)$.\newline
c)Cell body reference curve of minimum curvature deformation estimated after inversion.\newline
d)Cell body points registration along reference curve via dilation of $s_0(x,y)$ at a scale of $\sigma(x,y)$ and transformation of the resulting registration to elliptic coordinates.}
\label{fig13}
\end{figure}
\begin{figure}
\centering
\subfloat[]{\includegraphics[width=1.5in]{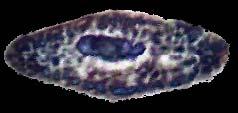}
\label{fig14a}}
\hfil
\subfloat[]{\includegraphics[width=1.25in]{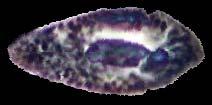}
\label{fig14b}}\\
\subfloat[]{\includegraphics[width=1.5in]{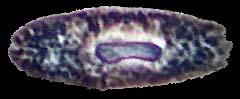}
\label{fig14c}}
\hfil
\subfloat[]{\includegraphics[width=1.35in]{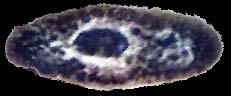}
\label{fig14d}}
\caption{Undeformed versions of the protozoa of Fig. \ref{fig12}, presented in the same order with the segmented images.}
\label{fig14}
\end{figure}
 
 The case of lips deformation is, in a sense, closer to the cells case. In fact, it is quite logical to adopt the assumption that there is a reference line of minimum curvature in the undeformed lips state and as a consequence we may apply the method of Sect. \ref{sec5_1} above. We, once more, note that all points of the undeformed body are expressed in elliptical coordinates with respect to this reference line. The validity of the aforementioned assumptions is supported by the following figures.
\begin{figure}[!t]
\centering
\subfloat[]{\includegraphics[width=1.5in]{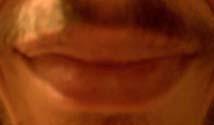}
\label{fig15a}}
\hfil
\subfloat[]{\includegraphics[width=1.5in]{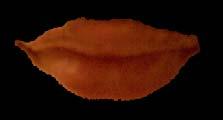}
\label{fig15b}}\\
\subfloat[]{\includegraphics[width=1.5in]{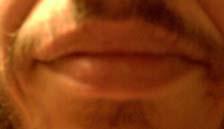}
\label{fig15c}}
\hfil
\subfloat[]{\includegraphics[width=1.5in]{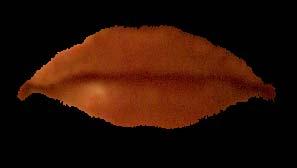}
\label{fig15d}}
\caption{Cropped and segmented human lips images of two different expressions of the same person.}
\label{fig15}
\end{figure}
\begin{figure}
\centering
\subfloat[]{\includegraphics[width=1in]{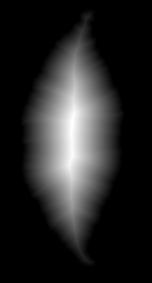}
\label{fig16a}}
\hfil
\subfloat[]{\includegraphics[width=.95in]{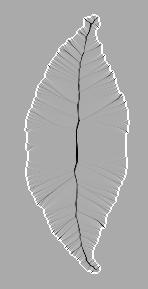}
\label{fig16b}}\\
\subfloat[]{\includegraphics[width=1in]{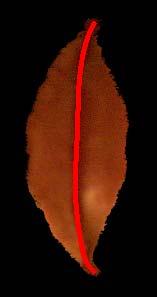}
\label{fig16c}}
\hfil
\subfloat[]{\includegraphics[width=.95in]{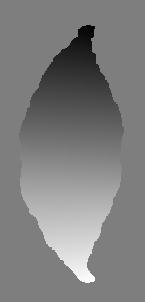}
\label{fig16d}}
\caption{Intermediate steps of the deformation inversion process of Sect. \ref{sec5_1} applied to the lips expression of Figs. \ref{fig15c}, \ref{fig15d}.\newline
a)Initialization of the filtering process; image, where the lips pixels have intensity $\delta_0(x,y)$.\newline
b)Image of the lips curvature deformation minimized at a scale $\sigma(x,y)$.\newline
c)Lips reference curve of minimum curvature deformation estimated after inversion.\newline
d)Lips points registration along reference curve via dilation of $s_0(x,y)$ at a scale of $\sigma(x,y)$ and transformation of the resulting registration to elliptic coordinates.}
\label{fig16}
\end{figure}
\begin{figure}
\centering
\subfloat[]{\includegraphics[width=1.5in]{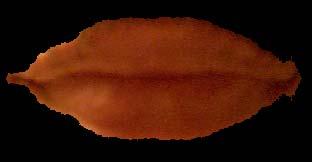}
\label{fig17a}}
\hfil
\subfloat[]{\includegraphics[width=1.5in]{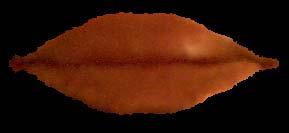}
\label{fig17b}}
\caption{Versions of the undeformed shape for the two different lips images of Fig. \ref{fig15}. It is evident that the introduced method, applied to the two different expressions furnished very similar undeformed states of the lips}
\label{fig17}
\end{figure}
\section{Automatic Classification of Deformed Objects - Parasites}
\label{sec6}
\subsection{Parasite Samples Acquisition}
\label{sec6_1}
  Diagnosis of the presence of parasitic infections in domestic animals, requires a coprological examination,   where eggs of the parasites strongyle family need to be   identified. Automatic identification of these parasites essentially reduces economic losses due to massive infection of domestic animals and it helps rapid restriction of the disease. The veterinary expert obtained parasite images from cultured faecal samples collected from naturally infected grazing sheep. According to the expert parasitologist, three to six images of each individual motile live larva, at 10x objective magnifications, were recorded by using a suitable charged coupled device camera mounted on the light microscope. A total of 317 images of 82 individual larvae in various shapes belonging to the genera Trichostrongylus(75 instances), Oesophagostomum (50 instances), Cooperia(35 instances), Ostertagia(33 instances), Haemonchus(52 instances), Teladorsasia(72 instances) were recorded. Each image was labeled with the genus name of the corresponding larva, its magnification and an identification number.
\subsection{The developed method for automatic classification of parasite contours}
\label{sec6_2}
We, once more, emphasize that the unwrapping process does not take into consideration at all that the deformed objects are parasites. Similarly, the automatic classification method presented below is very well applicable to any straightened object contours. In both cases the only restriction is the validity of the general assumptions made in Sects. \ref{sec3} and \ref{sec4}.
                                                                                                    We divide the available set of unwrapped parasite curves into a Training and a Test Set almost equal in number, by means of a random number generator. We also enumerate the individuals of the Training Set randomly. The expert, e.g. the parasitologist, classified all items of the Training Set into a number of groups based on the information the unwrapped parasites furnished. In order to test the efficiency of the classification method with maximum possible reliability, we have randomly generated a large number of pairs of Training and Test Set, say 200, and we have applied the classification method described below for each such pair separated.
                                                                                                    \subsubsection{The developed method for automatic classification of parasite contours}
\label{sec6_2_1}
First, we arrange all straightened contours of a group as follows:
We consider the unwrapped interpolated curve of the arbitrary i-th element of group  k, generated by means of the method described in Section \ref{sec4}: $\vec c^i_k(s)=\left(x^i_k(s) , y^i_k(s) \right)- \vec \mu_k(s) = \gamma^i_k(s) \hat n_k(s)$, where $\vec \mu_k(s)$ the symmetry line anf $\hat n_k(s)$ the unit vector normal to its tangents $\hat l_k(s)$.

We choose the contour of maximum length, say of group k with object index M, and we fit the unwrapped contours of the other samples of the group to it, via the following process:

Let the contour of maximum length be $\vec c_k^M(s )$ with the independent parameter of the length of the corresponding symmetry line $s \in [0,L_k^M]$. Then, the other objects' contours move along group's symmetry line so that their Euclidean distance from $\vec c_k^M(s)$ is minimal. Namely, for the $i$-th sample of group $k$, $\vec c_k^i(s)$, $s \in [0,L_k^i]$, we choose the translation $d^i_k$, $0 \leq d^i_k \leq L_k^M - L_k^i$ along group's symmetry line that minimizes the Euclidean distance $\sum_{s=0}^{L_k^i}\left\| \vec c_k^i(s) - \vec c_k^m(s+d^i_k) \right\|^2$ or equivalently maximizes the quantity 
$\sum_{s=0}^{L_k^i} \vec c_k^i(s) \cdot \vec c_k^m(s+d^i_k)$. Then we define the curve that represents $k$-group's object contours to be the mean curve of group's
 elements: $\vec c_k(s)= \frac{1}{N_k}\sum_{i=1}^{N_k(s)}\vec c_k^i(s-d^i_k)$, $s \in [0,L_k^M]$, $N_k(s)$ the number of $k$-group elements with translation $d^i_k \geq s$, $N_k$ the number of the elements of $k$-group.
\begin{figure*}
\centering
\subfloat[]{\includegraphics[width=3in]{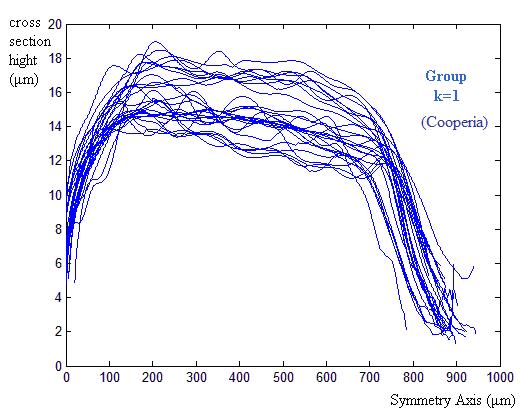}
\label{fig18a}}
\hfil
\subfloat[]{\includegraphics[width=3.3in]{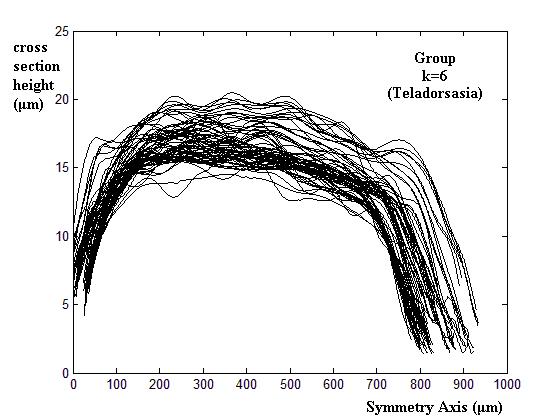}
\label{fig18b}}
\caption{ Two groups' Training Set parasite contours $\vec c_k^i(s)$ plotted per group (denoted with $k$) and centered around each group's contour of maximum length $\vec c_k^M(s)$}
\label{fig18}
\end{figure*}
\subsubsection{Tuning of the group representative curves of the Training Set}
\label{sec{6_2_2}}
We test the representability of the mean curves generated above, for the elements of the Training Set. In fact, we first attribute each element of the Training Set, whose curve is $\vec c_j(s)$, to the group with mean curve $\vec c_{k^*}(s)$, determined by the requirement that Euclidean distance of the two curves, computed as in Sect. \ref{sec6_2_1}, is minimum.

  When this attribution is over, we check its validity, taking into account the parasitologist's classification for the Training Set. If and only if a misclassification has occurred, we re-estimate the corresponding groups' representative curves as follows:

Let $\vec c^j_{w}(s)$ be the representative curve if the group to which $\vec c^j(s)$ has been erroneously attributed, while let $\vec c^j_{r}(s)$ be the representative curve of the group to which $\vec c^j(s)$ should have been attributed. Then, we first optimally match $\vec c^j(s)$ to $\vec c^j_{w}(s)$ as in Sect. \ref{sec6_2_1} and after that we re-estimate $\vec c^j_{w}(s)$ via the formula
\begin{eqnarray}
\vec c^j_{w}(s) &\leftarrow& \frac{N^j_{w}\vec c^j_{w}(s) - \beta^j_{w} \vec c^j(s+d^j_{w})}{N^j_{w} - \beta^j_{w}} , ~s \in [0,L^j_{w}] \nonumber \\
\beta^j_{w}&=&\exp \left(\frac{-1}{L^j_{w}+1}\sum_{s=0}^{L^j_{w}}\left\|\vec c^j_{w}(s)- \vec c^j(s+d^j_{w}) \right\|^2 \right) \nonumber \\
N^j_{w} &\leftarrow& N^j_{w} - \beta^j_{w}
\end{eqnarray}

Similarly, we let $\vec c^j(s)$ optimally match to $\vec c^j_r(s)$ and we re-estimate $\vec c^j_r(s)$ by means of the formula
\begin{eqnarray}
\vec c^j_{r}(s) &\leftarrow& \frac{N^j_{r}\vec c^j_{r}(s) + \beta^j_{r} \vec c^j(s+d^j_{r})}{N^j_{r} - \beta^j_{r}} , ~s \in [0,L^j_{r}] \nonumber \\
\beta^j_{r}&=&\exp \left(\frac{-1}{L^j_{r}+1}\sum_{s=0}^{L^j_{r}}\left\|\vec c^j_{r}(s)- \vec c^j(s+d^j_{r}) \right\|^2 \right) \nonumber \\
N^j_{r} &\leftarrow& N^j_{r} - \beta^j_{r}
\end{eqnarray}
  In both these re-estimation formulae quantities $\beta^j_{w}$ and $\beta^j_{r}$ are chosen so as to "punish" large distances between element and group curves and at the same time to support small distances.
  
We apply this process to all erroneously attributed elements of the Training Set, thus obtaining a new ensemble $E_2^{GR}$ of representative curves for each group.

  We repeat the above process by attributing all curves $\vec c^j(s)$ of the Training Set elements to the representative curves of group $E_2^{GR}$ and we check if the resulting classification has a success rate above a certain satisfactory threshold, say 98\%. If it does we stop the process and we consider the elements of $E_2^{GR}$ as actual representative curves of the existing curves and genera in the considered Training Set. Otherwise we repeat the previously described re-estimation for the curves of $E_2^{GR}$, thus creating a new ensemble $E_3^{GR}$ of representative curves for each group and so on, until the desired high success ratio is achieved. In the end of this process, an ensemble $E^GR$ of curves $\vec c_{k^*}(s)$ is obtained which represent the groups of the training set optimally.
\begin{figure*}
\centering
\subfloat[Plot of all ensembles $E_n^{GR}$ of groups k=1 and k=6 representative curves]{\includegraphics[width=3in]{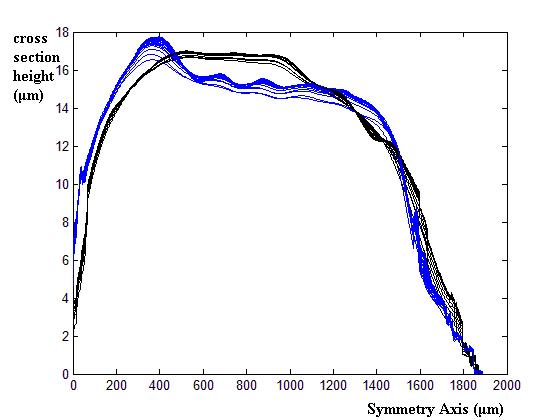}
\label{fig19a}}
\hfil
\subfloat[The final representative curves of these groups, in $E^{GR}$, that offer success rate over 98\%.]{\includegraphics[width=3in]{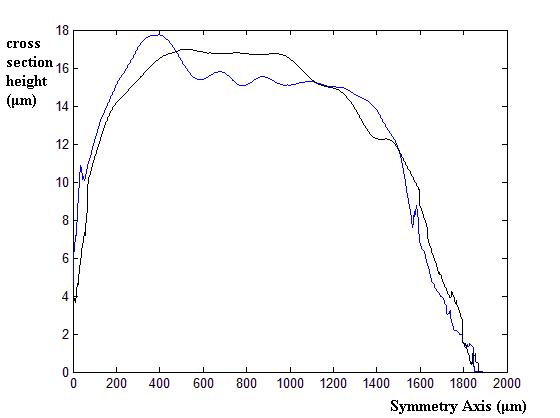}
\label{fig19b}}
\caption{Training process for the groups k=1 and k=6 representative curves}
\label{fig19}
\end{figure*}
\subsubsection{Classification of the elements of the Test Set}
\label{6_2_3}
  After generating $E^{GR}$, we proceed in the classification of the Test Set parasite images with a method analogous to the one described in the previous paragraph. More specifically, in connection with each available parasite image of the Test Set, we unwrap the deformed parasite by the method introduced in Sect. \ref{sec4}, forming a set of straightened parasite interpolated contours, say $C^{TS}$. Next, we compute the distance of each curve $\vec c^j(s)$ belonging in $C^{TS}$, from each group representative curve of ensemble $E^{GR}$ via the formula
\begin{equation}
\Delta(\vec c^j , \vec c_k) = \sum_{s=0}^{L^j}\left\| \vec c^j(s) - \vec c_k(s+d^j_k) \right\|^2 , ~\vec c_k(s) \in E^{GR}
\end{equation}
where $d_k^j$ is the translation along $\vec c_k(s)$ that centers $\vec c^j(s)$ and $\vec c_k(s)$ optimally in the least squares sense. Then, we attribute $\vec c^j(s)$ and the corresponding parasite to the group of minimum distance; namely $j$-th element of Test Set belongs to the $m$-th group if $m= \underset{k}{\arg \min}\left\{\Delta(\vec c^j , \vec c_k) \right\}$.
\section{Evaluation of the Introduced Methodology}
\label{sec7}
\subsection{Evaluation of the method of unwrapping the deformed bodies}
\label{sec7_1}
If the assumptions made in this paper and the introduced methodology are correct, one expects that different instances of a specific body deformation will generate very similar undeformed body shapes. In particular, one expects that different images of the deformed parasite must offer quiet close unwrapped versions after application of the methodology introduced in Sections \ref{sec4} and \ref{sec5}. In fact, Figures \ref{fig5A}, \ref{fig5B} demonstrate that this is indeed the case: The undeformed parasite borders have a difference that might be considered negligible in respect to the parasite dimensions. We have employed five different measures to describe the differences between the shapes of the unwrapped parasite that resulted from different phases. These measures are:
$a_{1,i}=\frac{l_i^P - \bar l^P}{\bar l^P} 100 \%$, where $l^P_i$ is the length of the unwrapped parasite obtained from the $i$-th wrapped larva phase and $\bar l^P$ its mean value, $a_{2,i}=\frac{E_i^P - \bar E^P}{\bar E^P} 100 \%$, where $E^P_i$ is the area of the unwrapped parasite obtained from the $i$-th wrapped larva phase and $\bar E^P$ its mean value,
$a_{3,i}=\frac{\underset{j}{\mbox{mean}}\left(y_{i,j} - \bar y_j \right)}{\underset{j}{\mbox{mean}}\left(\bar y_j \right)} 100 \%$, where $y_{i,j}$ is the width of the unwrapped parasite that corresponds to the $i$-th phase at point $x_j$; hence, we define $\bar y_j = \underset{i}{\mbox{mean}}(y_{i,j})$,
$a_{4,i}=\frac{\Pi_i^P - \bar \Pi^P}{\bar \Pi^P} 100 \%$, where $\Pi^P_i$ is the perimeter of the unwrapped parasite obtained from the $i$-th wrapped larva phase and $\bar \Pi^P$ its mean value,
$a_{5,i}=\frac{C_i^P - \bar C^P}{\bar C^P} 100 \%$, where $C^P_i$ is the maximum cross section diameter of the unwrapped parasite obtained from the $i$-th wrapped larva phase and $\bar C^P$ its mean value.

The mean value and standard deviation of quantities $a_{1,i}$, $a_{2,i}$, $a_{3,i}$, $a_{4,i}$, $a_{5,i}$ are shown in Table \ref{table1}.

Very similar undeformed shapes have been obtained, when the introduced methods have been applied to the images of deformed elastic fibers, protozoa and lips expressions. This is shown in Figs. \ref{fig11}, \ref{fig14} and \ref{fig17}. Evidently, the introduced method for deducing the undeformed body version from its deformation images, seems consistent, reliable and robust.
\begin{table*}[!t]
\renewcommand{\arraystretch}{1.3}
\caption{Parameters to Measure Consistency of the Unwrapped Parasite Contours }
\label{table1}
\centering
\begin{tabular}{|c||c||c||c|}
\hline
 & Evaluation of the discrepancy &	Evaluation of the discrepancy & Average discrepancy of all\\ & between the unwrapped & between the unwrapped & experimental results\\ & versions of Figure \ref{fig5A} &	versions of Figure \ref{fig5B} & \\ 
\hline
 & \begin{tabular}{c c} Mean &	Standard \end{tabular} & \begin{tabular}{c c} Mean &	Standard \end{tabular} & \begin{tabular}{c c} Mean &	Standard \end{tabular} \\
 & \begin{tabular}{c c} value \% & Deviation \% \end{tabular} & \begin{tabular}{c c} value \% &	Deviation \%  \end{tabular} & \begin{tabular}{c c} value \% &	Deviation \% \end{tabular} \\
\hline
$a_1$(Length) & \begin{tabular}{c||c} 0.82 & 0.29  \end{tabular} & \begin{tabular}{c||c} 1.04 &	0.73  \end{tabular} & \begin{tabular}{c||c} 0.90 &	0.61 \end{tabular} \\
\hline
$a_2$(Area) & \begin{tabular}{c||c} 2.30 &	1.18   \end{tabular} & \begin{tabular}{c||c} 1.41 &	0.62  \end{tabular} & \begin{tabular}{c||c} 1.67 &	1.04 \end{tabular} \\
\hline
$a_3$(Width) & \begin{tabular}{c||c} 1.87 &	1.06  \end{tabular} & \begin{tabular}{c||c} 1.32 &	0.60  \end{tabular} & \begin{tabular}{c||c} 1.69 &	0.91 \end{tabular} \\
\hline
$a_4$(Perimeter) & \begin{tabular}{c||c} 0.84 &	0.34   \end{tabular} & \begin{tabular}{c||c} 1.07 &	0.74  \end{tabular} & \begin{tabular}{c||c} 0.98 &	0.61 \end{tabular} \\
\hline
$a_5$(Max cross section) & \begin{tabular}{c||c} 2.63 &	1.68 \end{tabular} & \begin{tabular}{c||c} 2.62 &	1.63  \end{tabular} & \begin{tabular}{c||c} 2.68 &	1.70 \end{tabular} \\
\hline
\end{tabular}
\end{table*}
\subsection{Evaluation of the method of parasites automatic classification}
\label{sec7_2}
As described in Section \ref{sec6_1}, the entire set of data consisted of 317 parasite images that have been shot at an arbitrary deformation instance. We have randomly divided this set into 200 pairs of Training and Test Sets almost equal in number. For each Training Set we have applied the procedure described in Section \ref{sec4}, thus classifying all its members into groups, which in turn belong to one of the 6 genera.

	Next, we have considered all samples of the corresponding Test Set and we have automatically classified them to a corresponding group and gender. We have repeated the aforementioned process for all 200 pairs of Training and Test Sets and we have kept record of the correct and erroneous test classification of the Test Group in each case. The related results are summarized in Table \ref{table2}.
\begin{table}[!t]
\renewcommand{\arraystretch}{1.3}
\caption{Parasite Classification Results for All 200 Randomly Generated Test Sets.}
\label{table2}
\centering
\begin{tabular}{|c||c||c||c|}
\hline
Family & Number & Percentage of & Percentage of \\ & & Test Sets with & Test Sets with \\ & & Classifications & Classifications \\ & & of 0 erroneous & of 1 erroneous \\ & & identifications & identifications \\
\hline
1)Cooperia & 17 &	53\% & 41\% \\
\hline
2)Oesopha- & 24 & 30\% &	50\% \\ gostomum & & & \\
\hline
3)Ostertagia &	16 &	100\% &	0\% \\
\hline
4)Tricho- &	31 &	32\% &	48\% \\ strongylus & & & \\
\hline
5) Haemonchus &	21 &	29\% &	71\% \\
\hline
6) Teladorsasia &	29 &	52\% &	44\% \\
\hline
\end{tabular}
\end{table}
\subsection{Some aspects concerning the efficiency of the introduced methods}
\label{sec7_3}
The efficiency of the methods presented here, developed for constructing undeformed body shapes and images, relies also in the accuracy of segmentation and body extraction results. Once the body shape has been extracted from the image of its deformation instance with noisy but non-conceptual distortion (zero mean value noise), both methods developed for constructing the unwrapped body version can circumvent this inaccuracy. Namely, for both methods described in Sect. \ref{sec4_4}, \ref{sec4_5} and in Sect. \ref{sec5}, local distortion of the body contour does not affect the polynomial form optimally fitted to it, since we expect zero mean value for this local distortion and the polynomial fits the contour points in the least squares sense. The method employed here for segmentation of parasite images makes use of the color intensity local homogeneity under the assumption that this homogeneity is represented by a normal distribution of color intensity and hence local inaccuracies have zero mean value. So, provided that this method offered reliable extraction of body shape, local inaccuracies do not affect the introduced unwrapping methods' performance. In any case, if in another application this segmentation method couldn't offer reliable body shapes, different segmentation approaches (active contours \cite{bib6}, watershed-based algorithms \cite{bib11}, etc.) could be employed.

Another aspect about the efficiency of the methods introduced here, concerns their applicability to the reconstruction of body undeformed shape in other entities than parasites. In the sense that the considered bodies suffer an equal to 2D deformation, analysis of Sect. \ref{sec3_3} is valid. But the reliable estimation of the body unwrapped version depends on the demand that body deformation is invertible and on the reliability of the estimation of body undeformed reference curve. Namely, if the assumptions described in Sect. \ref{sec3_1} hold for the considered body then the performed experiments show that both methods described in Sect. \ref{sec4_4}, \ref{sec4_5} and in Sect. \ref{sec5_1} are directly applicable provided that there is a symmetry line in the undeformed body state. This is the case for the elastic deformation of various objects/entities, such as fibers, cables, nails, sheets, but also serpents and eels, etc. On the other hand, there is a variety of entities that may suffer elastic deformation of the form described in Sect. \ref{sec3_1} which in their undeformed state have a piecewise smooth curve of minimum curvature, i.e. the reference curve. Such bodies are cells, viruses, the human lips, etc. One may, also, tackle the problem of retrieving the undeformed shape of such a body by applying the method described in Sect. \ref{sec5_1}. Characteristic examples of the aforementioned cases, additional to the parasite one, are given in Sect. \ref{sec5_2} supporting the applicability of the introduced methods. We would like to emphasize that the introduced methods are valid both for rectilinear and curvilinear coordinates describing the body points.

Concerning the developed curve classification method, the efficiency of its classification performance depends on how representative are the mean curves of each group for the curves that belong to it. In these cases of bodies, such as parasites, cells, fibers, sheets, industrial line's objects, etc., undeformed shapes of objects of the same kind are very close to each other differing only randomly from an ideal shape representative for the object kind. Hence, in such cases the mean contour is indeed a reliable representative of the undeformed body shapes of objects of the same kind. In this sense, the classification problem encountered in the present work is not a trivial one. All considered "objects" are parasites of the same kind and, as Figures \ref{fig18}, \ref{fig19} indicate, their undeformed shape contours and the class representative mean curves are very close to each other making parasite family separation a difficult task. So, in order to obtain a high parasite family identification rate, we have developed a tuning process for the family representative curves in order to increase the separability between the corresponding groups.

\section{Conclusion}
In this paper, a new methodology has been introduced that tackled two goals:\\
1. To exploit images of elastic body deformation instances, so as to verify assumptions about its mechano-elastic properties. The validity of these assumptions allows for unwrapping the body, i.e. for obtaining the body undeformed version from its deformed image. This has been achieved here a) by detecting elastic deformation invariant curves (neutral line and its cross sections), b) by applying to the image of a 2D body deformation the inverse deformation process, as it is constructed by a sequence of image operations equivalent to solving the differential equations governing the body deformation.\\
2. To automatically classify the deformed bodies (here parasites) on the basis of comparison of the contours of their undeformed version. To achieve this, a curve classification method has been developed that creates a representative curve for each group of the Training Set and classifies each element of a Test Set according to the distance of the unwrapped contour of this element from all these representative curves.
Application of this methodology to 317 images of highly deformed parasites of domestic animals, offered straightened contours and body versions that seem to be consistent and reliable representations of the undeformed parasites. In addition, we have obtained the undeformed shape of fibers, cells and human lips by application of the methodology introduced here, so as to make its applicability to various entities more clear. 

Concerning the problem of deformed parasites identification, which was the motivation of the present work, we would like to point out that the expert parasitologist could classify the gene from the obtained parasite unwrapped versions. However, the authors have employed these unwrapped parasite versions developing an automatic classification system for the parasite genera. The classification method developed by the authors, succeeds in classifying the deformed parasites to proper groups and six families with more than 97.6 \% success rate for 200 randomly chosen combinations of Training and Test Sets.

%

\appendices
\section{}
Here, we analytically construct the solution of the PDE problem
\begin{equation}
\frac{\partial}{\partial s}\ln \tilde w = -\frac{\partial}{\partial n}\left(\ln \left\| \nabla \tilde w \right\| \right)
\label{A1}
\end{equation}
with initial conditions $\left\{\tilde w(x,y) \right\}(0,\delta)=M^{+}_{\delta}\left[\delta_0(x,y) \right]$ and $\left\{\tilde \phi_0(x,y) \right\}(0,\delta)=\left(\ln \left\| \nabla \tilde w \right\| \right)(0,\delta)$. For solving it, first we express the directional derivative of $\tilde w_n$ at an arbitrary direction $\hat v$,
\begin{equation}
\tilde w_{nv} = \hat v^T H(\tilde w) \hat n =\tilde w_{nl}\hat v^T \hat l + \tilde w_{nn}\hat v^T \hat n 
\label{A2}
\end{equation}
But as it has been mentioned in Sect. \ref{sec3_3} curvature is related with functional $\tilde w$ via the relation (\ref{3_3_2}). Performing dot product of (\ref{3_3_2}) with $\hat n$, there results $\frac{\tilde w_{nl}-\hat n^T \hat l \tilde w_{nn}}{\left\| \nabla \tilde w \right\|} = \kappa \hat n^T \hat l \Rightarrow \tilde w_{nl} =0$. By substituting in (\ref{A2}) we obtain $\tilde w_{nv} = \tilde w_{nn} \hat v^T \hat n$. This relation bounds $\tilde w_{nv}$ and, consequently, $\frac{\tilde w_{nv}}{\left\| \nabla \tilde w \right\|}=\frac{\partial}{\partial v}\left(\ln \left\| \nabla \tilde w \right\| \right)$. Writing $-\frac{\partial}{\partial n}\left(\ln \left\| \nabla \tilde w \right\| \right)$ as an extreme of $\frac{\partial}{\partial v}\left(\ln \left\| \nabla \tilde w \right\| \right)$ one obtains:\\
For $s : \kappa(s) <0$, \\$-\frac{\partial}{\partial n}\left(\ln \left\| \nabla \tilde w \right\| \right) = \frac{1}{ds}\underset{\left\| \vec v \right\|\leq ds}{\sup}\left\{\vec v^T \nabla \left(\ln \left\| \nabla \tilde w \right\| \right) \right\}=$\\$=\underset{ds \to 0}{\lim}{\frac{\underset{\left\| \vec v \right\|\leq ds}{\sup}\left[\left\{\ln \left\| \nabla \tilde w \right\| \left((x,y)+ \vec v \right) \right\}(s,\delta) \right]-\left\{\ln \left\| \nabla \tilde w \right\| (x,y) \right\}(s,\delta)}{ds}}=$\\
$=\frac{\partial}{\partial s}\underset{\left\| \vec v \right\|\leq s}{\sup}\left[\left\{\ln \left\| \nabla \tilde w \right\| \left((x,y)+ \vec v \right) \right\}(0,\delta) \right]=\frac{\partial}{\partial s}M^{+}_{s}\left[\left\{\tilde \phi_0(x,y) \right\}(0,\delta) \right]$, i.e. the derivative of the dilation of $\left\{\tilde \phi_0(x,y) \right\}(0,\delta)=\left\{\ln \left\| \nabla \tilde w \right\|(x,y) \right\}(0,\delta)$ at scale of $s$.\\
For $s : \kappa(s) >0$, \\$-\frac{\partial}{\partial n}\left(\ln \left\| \nabla \tilde w \right\| \right) = \frac{1}{ds}\underset{\left\| \vec v \right\|\leq ds}{\inf}\left\{\vec v^T \nabla \left(\ln \left\| \nabla \tilde w \right\| \right) \right\}=$\\$=\underset{ds \to 0}{\lim}{\frac{\underset{\left\| \vec v \right\|\leq ds}{\inf}\left[\left\{\ln \left\| \nabla \tilde w \right\| \left((x,y)+ \vec v \right) \right\}(s,\delta) \right]-\left\{\ln \left\| \nabla \tilde w \right\| (x,y) \right\}(s,\delta)}{ds}}=$\\
$=\frac{\partial}{\partial s}\underset{\left\| \vec v \right\|\leq s}{\inf}\left[\left\{\ln \left\| \nabla \tilde w \right\| \left((x,y)+ \vec v \right) \right\}(0,\delta) \right]=\frac{\partial}{\partial s}M^{-}_{s}\left[\left\{\tilde \phi_0(x,y) \right\}(0,\delta) \right]$, i.e. the derivative of the erosion of $\left\{\tilde \phi_0(x,y) \right\}(0,\delta)=\left\{\ln \left\| \nabla \tilde w \right\|(x,y) \right\}(0,\delta)$ at scale of $s$.

Then (\ref{A1}) implies 
\begin{equation}
\frac{\partial}{\partial s}\ln \tilde w=\left\{ \begin{array}{ll}
\frac{\partial}{\partial s}M^{+}_{s}\left[\left\{\tilde \phi_0(x,y) \right\}(0,\delta) \right] & \kappa(s_0(\delta))<0 \\
\frac{\partial}{\partial s}M^{-}_{s}\left[\left\{\tilde \phi_0(x,y) \right\}(0,\delta) \right] & \kappa(s_0(\delta))>0 
\end{array}
\right.
\label{A3}
\end{equation}
where $s_0(\delta)$ is given for $\delta$-deformation using (\ref{3_4_2}) and the initial condition $\left\{\tilde \phi_0(x,y) \right\}(0,\delta)$ can be derived immediately from a dilated version of the global initial conditions $(s_0,\delta_0)$. To unify both these cases in one process we define the function $\left\{\kappa_{\phi}(x,y) \right\}(s_0(\delta)) = -\mbox{sgn}\left(\kappa(s_0(\delta)) \right)\infty$.

We adopt the process
\begin{equation}
\alpha[g(\delta)](\sigma) = \sup \left[M^{-}_{\sigma}[g(\delta)] , \inf \left[M^{+}_{\sigma}[g(\delta)] , \kappa_{\phi}(s_0(\delta)) \right] \right]
\label{A4}
\end{equation}
The derivative of this functional reads
\begin{equation}
\frac{\partial}{\partial \sigma}\alpha[g(\delta)](\sigma) = \left\{
\begin{array}{ll}
\frac{\partial}{\partial \sigma}M^{+}_{\sigma}[g(\delta)] & \begin{array}{ll} \kappa_{\phi}(s_0(\delta))<M^{+}_{\sigma}[g(\delta)] \Leftrightarrow \\ \kappa(s_0(\delta))<0 \end{array} \\
\frac{\partial}{\partial \sigma}M^{-}_{\sigma}[g(\delta)] & \begin{array}{ll} \kappa_{\phi}(s_0(\delta))>M^{+}_{\sigma}[g(\delta)]\Leftrightarrow  \\ \kappa(s_0(\delta))>0 \end{array}
\end{array}\right.
\end{equation}

Letting $g(\delta) = \left\{\tilde \phi_0(x,y) \right\}(0, \delta)$ filter $\alpha$ on $s$-scale space gives a solution for (\ref{A3}) and equivalently of (\ref{A1}). Finally, correspondence between (\ref{A1}) with initial conditions $g(\delta)$ and filter $\alpha[g(\delta)](s)$ results $\left\{\tilde w(x,y)\right\}(s,\delta)=\left\{\tilde w(x,y)\right\}(0,\delta) \exp \left(\alpha \left[\left\{\tilde \phi_0(x,y)\right\}(0,\delta) \right](s) \right)$
\section{}
Let $\delta(s) = |\vec{MA}|=|\vec{MD}|$ in the undeformed body, namely the distance of an arbitrary point of the upper or the lower boundary curve from reference line expressed as a function of length s of reference line. We describe the upper boundary curve $U'$ of the deformed body by the vector equation $\vec r_{U'}(s)$, where the parameter $s$ is already defined to be the length of $\Sigma_1 M$; similarly, let the vector parametric equation for the lower boundary be $\vec r_{L'}(s)$. Therefore, taking (\ref{3_3_1}) into consideration we deduce that
\begin{equation}
d \vec r_{U'} = \left(\left\| \nabla \tilde u \right\| - \tilde w \frac{\tilde w_{nn}}{\left\| \nabla \tilde w \right\|} \right)ds \hat l(s) +ds \dot \delta(s) \left\| \nabla \tilde w \right\| \hat n(s)
\end{equation}
\begin{equation}
d \vec r_{L'} = \left(\left\| \nabla \tilde u \right\| + \tilde w \frac{\tilde w_{nn}}{\left\| \nabla \tilde w \right\|} \right)ds \hat l(s) -ds \dot \delta(s) \left\| \nabla \tilde w \right\| \hat n(s)
\end{equation}
where, as usual, $\hat l(s)$ is the unit vector tangent to the neutral line ata $M'$ (the image of $M$ in the considered deformation) and $\hat n(s)$ the unit vector normal to $\hat l(s)$ that points towards $U'$.

Next we compute the inner product of tangent vectors $\dot {\vec r}_{U'}$ and $\dot{\vec r}_{L'}$ with cross section $\vec{D'A'}$ thus obtaining
\begin{equation}
\left(\dot{\vec r}_{U'} - \left\| \nabla \tilde u \right\| \hat l(s) \right)\cdot \vec{D'A'} = \dot{\vec r}_{U'} \cdot \vec{D'A'}= \dot \delta(s) \left\| \nabla \tilde w \right\| 
\label{B3}
\end{equation}
\begin{equation}
\left(\dot{\vec r}_{L'} - \left\| \nabla \tilde u \right\| \hat l(s) \right)\cdot \vec{D'A'} = \dot{\vec r}_{L'} \cdot \vec{D'A'}= -\dot \delta(s) \left\| \nabla \tilde w \right\|
\label{B4}
\end{equation}
It also holds that 
\begin{equation}
\left|\frac{d}{ds}\vec r_{U'} - \left\| \nabla \tilde u \right\| \hat l(s) \right|=\left|\frac{d}{ds}\vec r_{L'} - \left\| \nabla \tilde u \right\| \hat l(s) \right|
\label{B5}
\end{equation}
But $\left\| \nabla \tilde u \right\| \hat l(s)$ is the tangent vector of the deformed symmetry line at length $s$, namely $\vec L_{\mu}(s)=\dot{\vec r}_{\mu}(s)=\left\| \nabla \tilde u \right\| \hat l(s)$. Now, if we let $\xi$ be the angle between $\dot{\vec r}_{U'}-\vec L_{\mu}(s)$ and $\vec{D'A'}$ and $\theta$ the one between $\dot{\vec r}_{L'} - \vec L_{\mu}(s)$ and $\vec{D'A'}$, then, combining (\ref{B3}), (\ref{B4}) and (\ref{B5}) we obtain $\frac{\dot{\vec r}_{U'}\cdot \vec{D'A'}}{\dot{\vec r}_{L'}\cdot \vec{D'A'}}= \frac{\cos \xi}{\cos \theta}=-1$ and, as a consequence, $\cos \xi = -\cos \theta$, which results $\xi + \theta = \pi$.
\ifCLASSOPTIONcaptionsoff
  \newpage
\fi


\bibliographystyle{IEEEtran}
%

%

\begin{IEEEbiography}[{\includegraphics[width=1in,height=1.25in,clip,keepaspectratio]{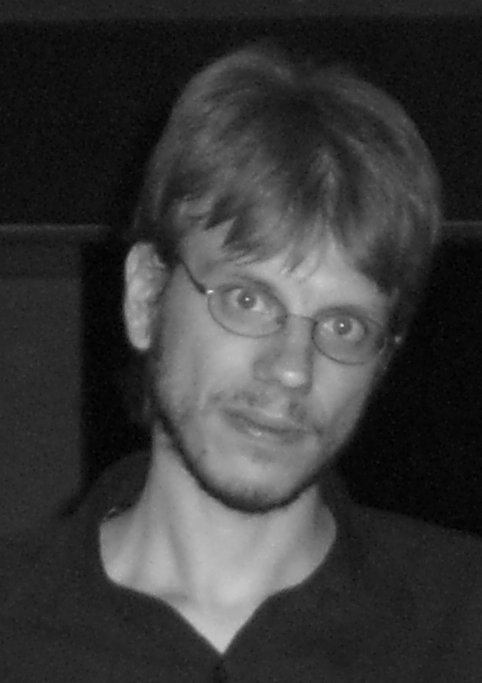}}]{Dimitris Arabadjis}
received the Diploma degree in electrical and computer engineering from the National Technical University of Athens in 2006. He has been a PhD student in the School of Electrical and Computer Engineering of National Technical University of Athens (NTUA) since 2007. His research interests involve the following subjects: pattern recognition, image processing, applied \& computational geometry, curve fitting methods, biomedical engineering, etc. He has three publications in international journals on these subjects.
\end{IEEEbiography}
\begin{IEEEbiography}[{\includegraphics[width=1in,height=1.25in,clip,keepaspectratio]{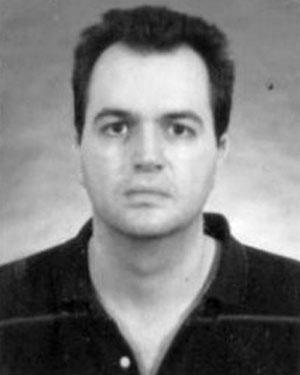}}]{Panayiotis Rousopoulos}
received the Diploma degree in physics from the University of Patras in 2002. He has been a PhD student in the School of Electrical and Computer Engineering of National Technical University of Athens (NTUA) since 2002. His research interests and recent work are on the following subjects: applications of information theory to archaeology, image processing, pattern recognition, finite precision error, numerical solutions of differential equations, etc. He has twelve publications in international journals on these subjects.
\end{IEEEbiography}
\begin{IEEEbiography}[{\includegraphics[width=1in,height=1.25in,clip,keepaspectratio]{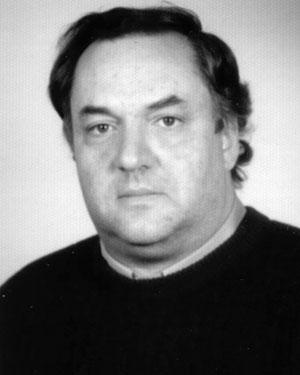}}]{Constantin Papaodysseus}
received the Diploma degree in electrical and computer engineering from the National Technical University of Athens (NTUA) and the MSc degree from Manchester University, United Kingdom. He received the PhD degree in computer engineering from NTUA. From 1996-2000, he was an assistant professor at NTUA in the Department of Electrical and Computer Engineering. Since 2001, he has been an associate professor in the same department of NTUA. His research interests include image processing, pattern recognitio, biomedical engineering, music and sound processing and automatic recognition, applications of computer science to archaeology, applied mathematics, algorithm robustness and quantization error analysis, adaptive algorithms, etc. He has more than 45 publications in international journals and many publications in international conferences on these subjects.
\end{IEEEbiography}
\begin{IEEEbiography}[{\includegraphics[width=1in,height=1.25in,clip,keepaspectratio]{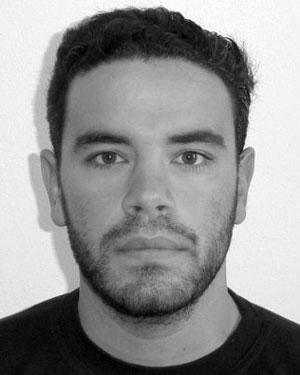}}]{Michalis Panagopoulos}
received the Diploma degree in electrical and computer engineering from the National Technical University of Athens in 2001. He recieved the PhD degree in computer engineering from the School of Electrical and Computer Engineering of the NTUA in 2008. His main research interests involve: Image processing, pattern recognition, curve fitting, finite precision error, biomedical engineering, and application of pattern recognition methods and statistics to archaeology. He has eight publications in international journals on these subjects.
\end{IEEEbiography}
\begin{IEEEbiography}[{\includegraphics[width=1in,height=1.25in,clip,keepaspectratio]{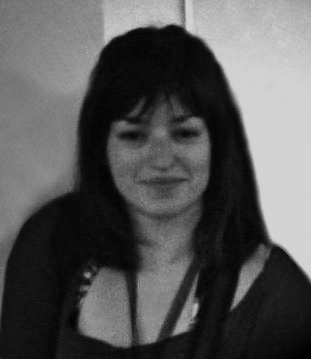}}]{Panayiota Loumou}
was born in Athens, Greece in 1985. She is a graduate student at the National Technical University of Athens, Department of Electrical \& Computer Engineering. She is a member of the Multimedia Laboratory of the same department and has participated in several European Programs as a Project Team Member. Her research interests involve biomedical engineering, image processing, web development and social networks.
\end{IEEEbiography}
\begin{IEEEbiography}[{\includegraphics[width=1in,height=1.25in,clip,keepaspectratio]{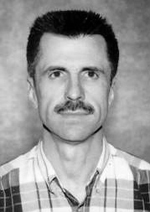}}]{George Theodoropoulos}
is an Associate Professor at the Agricultural University of Athens, Department of Anatomy \& Physiology of Domestic Animals. He received his Diploma of Veterinary Medicine from the Aristotle University of Thessaloniki (1979) and his M.Sc. degree in Veterinary Pathology (Veterinary Parasitology) from the Iowa State University (1984). He received his Ph.D. degree in Comparative Pathology, from the University of California, Davis (1988). 
His research interests include host-parasite interactions, epidemiology of parasites, multimedia computer systems in Veterinary Medicine, etc. He has more than forty-five publications in international Journals and many publications in international conferences on these subjects.
\end{IEEEbiography}






\end{document}